%% file: main.tex
\documentclass[10pt,letterpaper,conference]{ieeeconf}

\IEEEoverridecommandlockouts                              %
\usepackage{amsmath, amssymb}

\makeatletter
\let\NAT@parse\undefined
\makeatother
\usepackage{comment}
\let\labelindent\relax
\usepackage{enumitem} %
\usepackage{hyperref}
\hypersetup{
    colorlinks=true,
}
\usepackage{booktabs}
\usepackage{pdfpages}
\usepackage{multirow}
\usepackage{nicefrac}
\usepackage[normalem]{ulem}
\useunder{\uline}{\ul}{}
\usepackage{array}
\newcolumntype{P}[1]{>{\centering\arraybackslash}p{#1}}
\newcolumntype{M}[1]{>{\centering\arraybackslash}m{#1}}
\usepackage{pifont}%
\usepackage{adjustbox}
\usepackage{capt-of}
\usepackage{balance} %

\newcommand{\bez}{B\'{e}zier}
\newcommand{\rf}{\ensuremath{\text{r}}}
\newcommand{\tg}{\ensuremath{\text{t}}}
\newcommand{\crr}{\ensuremath{\text{CR}}}
\newcommand{\cnt}{\ensuremath{\text{CT}}}
\newcommand{\corrv}{\ensuremath{\mathbf{C}}}
\newcommand{\simdset}{MultiFlow}
\newcommand{\tldata}{HS-ERGB}

\usepackage{color} 
\newcommand{\changes}[1]{\textcolor{black}{#1}}
\newenvironment{changedtable}
  {\color{black}}
  {}

\definecolor{somegray}{rgb}{0.5, 0.5, 0.5}
\newcommand{\darkgrayed}[1]{\textcolor{somegray}{#1}}
\makeatletter
\newcommand*\titleheader[1]{\gdef\@titleheader{#1}}
\AtBeginDocument{%
  \let\st@red@title\@title
  \def\@title{%
    \vskip-3em
    \bgroup\normalfont\large\centering\@titleheader\par\egroup
    \vskip1.5em\st@red@title}
}
\makeatother

\titleheader{\darkgrayed{This paper has been accepted for publication at\\
IEEE Transactions on Pattern Analysis and Machine Intelligence (TPAMI), 2024.
\copyright IEEE}}

\title{\LARGE \bf
Dense Continuous-Time Optical Flow from Event Cameras
}

\author{Mathias~Gehrig,
        Manasi~Muglikar, and
        Davide~Scaramuzza%
\thanks{The authors are with the Robotics and Perception Group, affiliated with both the Dept. of Informatics of the University of Zurich and the Dept. of Neuroinformatics of the University of Zurich and ETH Zurich, Switzerland.
}%
}

\newcommand{\imgfigWidthX}{0.180\linewidth}
\newcommand{\thisfigWidthX}{0.180\linewidth}
\makeatletter
\let\@oldmaketitle\@maketitle%
\renewcommand{\@maketitle}{\@oldmaketitle%
\input{figs/eyecatch_page1}}
\makeatother

\begin{document}

\maketitle
\thispagestyle{plain}
\pagestyle{plain}

\input{sections/abstract.tex}

\input{sections/introduction_v2.tex}
\input{sections/relwork.tex}

\input{sections/methodology.tex}

\input{sections/dataset.tex}

\input{sections/experiments.tex}

\input{sections/discussion.tex}

\section*{Acknowledgment}
This work was supported by the National Centre of Competence in Research (NCCR) Robotics (grant agreement No. 51NF40-185543) through the Swiss National Science Foundation (SNSF), and the European Research Council (ERC) under grant agreement No. 864042 (AGILEFLIGHT).

\bibliographystyle{IEEEtran}
\bibliography{all}



\section*{Supplementary material}

# Supplementary Material

## I. INFLUENCE OF TRAJECTORY LENGTH ON THE TRAJECTORY ERROR

In Figure 1, we examine how the trajectory length influences the trajectory error on the MultiFlow. As expected, the trajectory error increases when the trajectory becomes longer because the task complexity increases. For example, longer trajectories correlate with more complex motion patterns and occlusions which are particularly challenging for optical flow methods.

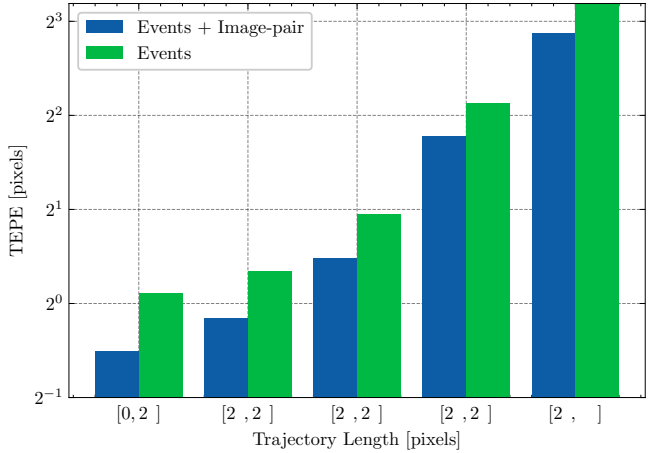

Fig. 1. Trajectory end-point-error (TEPE) as a function of the trajectory length. We use the $\log_2$ scale to better capture the full range of trajectory length and errors. The error increases with the trajectory length, and image data generally lowers the error.

## II. NETWORK ARCHITECTURE

Figure 2 shows the neural network architecture used throughout this paper. The architecture is based on the RAFT neural network [1] with modifications highlighted in **red**. We use the same encoder architecture to extract context features from voxel grids (events) and, optionally, the image at the reference timestamp, as well as correlation features from both modalities. This encoder architecture is identical to the RAFT context and correlation encoder. Instead, the update block requires a few modifications:

- Instead of encoding pixel displacements (flow), we encode the Bézier curve parameters $\mathbf{P}_{k-1}$ from the previous output of the update block.
- We add a context encoder to extract motion features directly from event data and optionally use the image context encoder.
- We predict $2n$ scalars per feature of the update block to estimate the increment of $n$ two-dimensional Bézier curve control points $\mathbf{\Delta P}_k$.

The upsampling module, omitted for brevity, is identical to the one used in RAFT.

## III. MULTIFLOW DATASET GENERATION

This section provides a more formal and detailed overview of the data generation process that was used to create the MultiFlow dataset.

The duration of each sequence in the dataset is fixed to one second. We sample either 3 or 4 (random choice of 50% probability each) control points in the space of pixel translation, scale, and rotation of the foreground (FG) and background (BG) objects in the image plane. The time assigned to control points is sampled from the range $[0, 1]$ while ensuring strictly increasing values to maintain the temporal order.

The control points are sampled according to the following discrete-time process with parameters shown in Table I:

$$X_{k+1} = \begin{cases} \hat{\gamma} \cdot \hat{X}_{k+1}^{\text{Det}} + (\hat{\gamma} - 1) \cdot \hat{X}_{k+1}^{\text{Stoch}} & (\hat{\alpha} = 0) \wedge (\hat{\beta} = 0) \\ X_k & (\hat{\alpha} = 1) \vee (\hat{\beta} = 1) \end{cases} \quad (1)$$

where

$$\hat{X}_{k+1}^{\text{Det}} = X_k + \frac{t_{k+1} - t_k}{t_k - t_{k-1}}(X_k - X_{k-1}) \quad (2)$$

refers to the deterministic part of the process that uses a constant velocity model. It is inspired by the law of conservation of momentum from classical mechanics. $t_k$ is the time associated with control point $k$. To introduce more variability in the data generation process, we introduce a stochastic process:

For *translation and rotation*:

$$\hat{X}_{k+1}^{\text{Stoch}} = X_k + \Delta X \quad \Delta X \sim \text{Uni}(-\theta, \theta) \quad (3)$$

For *scale*:

$$\hat{X}_{k+1}^{\text{Stoch}} = X_k \cdot \Delta X \quad \Delta X \sim (1 + \delta_0)^{(2\delta_1 - 1)}, \quad (4)$$
$$\delta_0 \sim \text{Uni}(0, \theta), \ \delta_1 \sim \text{Bern}(0.5)$$

*Uni* stands for the continuous uniform distribution and *Bern* refers to the Bernoulli distribution.

$\hat{\alpha} \sim \text{Bern}(\alpha)$ models the probability that the similarity transformation (translation, rotation *and* scale) of the object remains constant during the whole sequence.

1

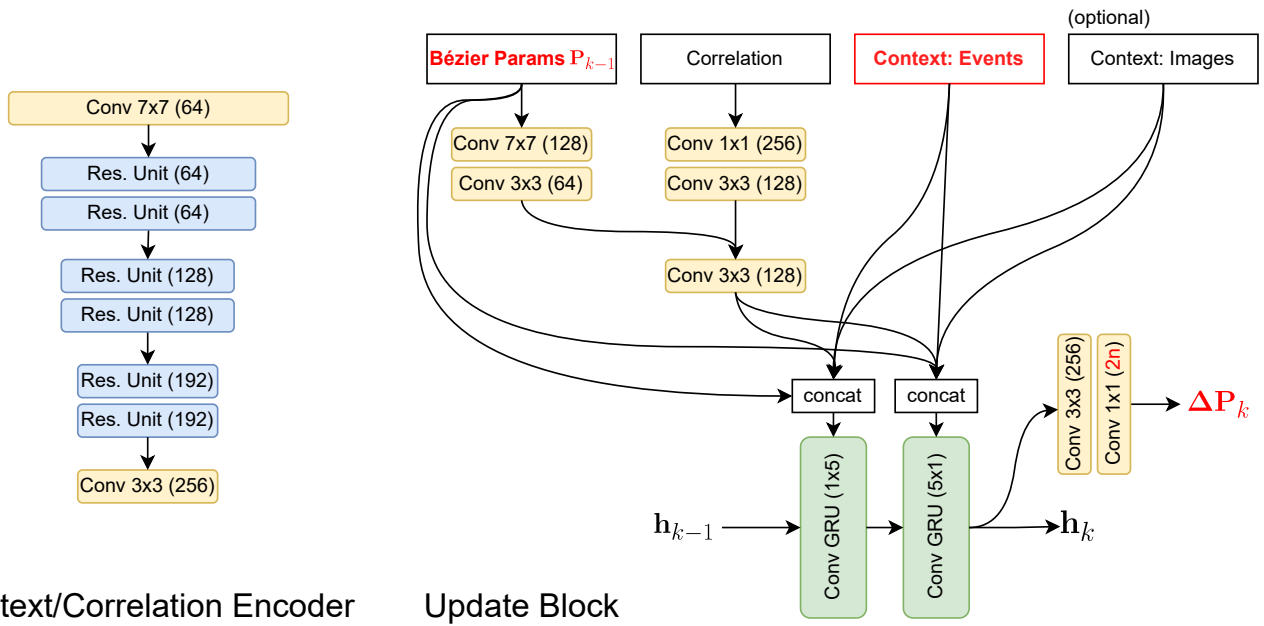

Fig. 2. Neural network architecture adapted from RAFT to predict Bézier curves. Modifications in the update block are highlighted in red. Context and correlation encoder architectures remain identical to RAFT, with an added context encoder for event data. $n$ refers to the degree of the Bézier curve.

- $\hat{\beta} \sim \text{Bern}(\beta)$ is sampled separately for each component (translation, rotation *or* scale) of the similarity transformation and models the probability that the component remains constant during the sequence.
- $\hat{\gamma} \sim \text{Uni}(0, \gamma)$ is sampled separately for each component (translation, rotation *or* scale) of the similarity transformation and models the degree of the constant velocity component. For $\hat{\gamma} = 1$, the new control point is computed deterministically with a constant velocity model according to equation (2). $\hat{\gamma} = 0$ corresponds to a purely stochastic update according to equations (3) or (4). $\hat{\gamma} \in (0, 1)$ results in a convex combination of both as shown in equation (1).

| Component | $\alpha$ | $\beta$ | $\gamma$ | $\theta$ |
|---|---|---|---|---|
| Translation BG | | 0 | 0.8 | 30 |
| Rotation BG | 0.1 | 0.7 | 0.6 | 10 |
| Scale BG | | 0.4 | 0.3 | 0.15 |
| Translation FG | | 0 | 0.9 | 120 |
| Rotation FG | 0 | 0.3 | 0.6 | 30 |
| Scale FG | | 0.3 | 0.3 | 0.30 |

TABLE I

PARAMETERS OF THE DISTRIBUTION OF THE TRANSFORMATION COEFFICIENTS FOR FOREGROUND OBJECTS (FG) AND BACKGROUND (BG). THE VALUES OF $\theta$ ARE SHOWN IN *pixels* FOR TRANSLATION, IN *degree* FOR ROTATION AND IN *dimensionless quantity* FOR SCALE. THE VALUES OF $\alpha$ AFFECT THE OVERALL SIMILARITY TRANSFORMATION AND THEREFORE BELONG TO TRANSLATION, ROTATION AND SCALE SIMULTANEOUSLY.

We compute continuous-time trajectories over similarity transformations via cubic spline interpolation on the $K \in \{3, 4\}$ acquired control points $X_0, \ldots, X_{K-1}$. It would be possible to sample a higher number of control points, but we found that 4 control points are sufficient to capture interesting non-linear motion that we could also observe in real-world applications of our algorithm (see Fig. 1 and 7 in the paper).

**Pixel Trajectory Ground Truth:** The ground truth for the pixel trajectories is computed with respect to a reference time at $0.4$ seconds. The pixel trajectory is expressed as pixel displacements at different target timestamps with respect to the original pixel location at the reference time. Overall, the dataset consists of pixel trajectory ground truth at intervals of 10 milliseconds up to time $0.9$ seconds. The first $0.4$ and last $0.1$ seconds can be used as additional context.

**Visual Examples:** Figure 3 shows examples of frames at reference timestamp in column (a) and target timestamp in column (b) as well as corresponding predictions of our model trained on the dataset. Frames between the reference and target timestamps are used to generate events that are used by our model to predict the continuous-time pixel trajectories. In column (c) we visualize the predicted pixel displacement between the reference and target timestamp in the Sintel colorization scheme [2].

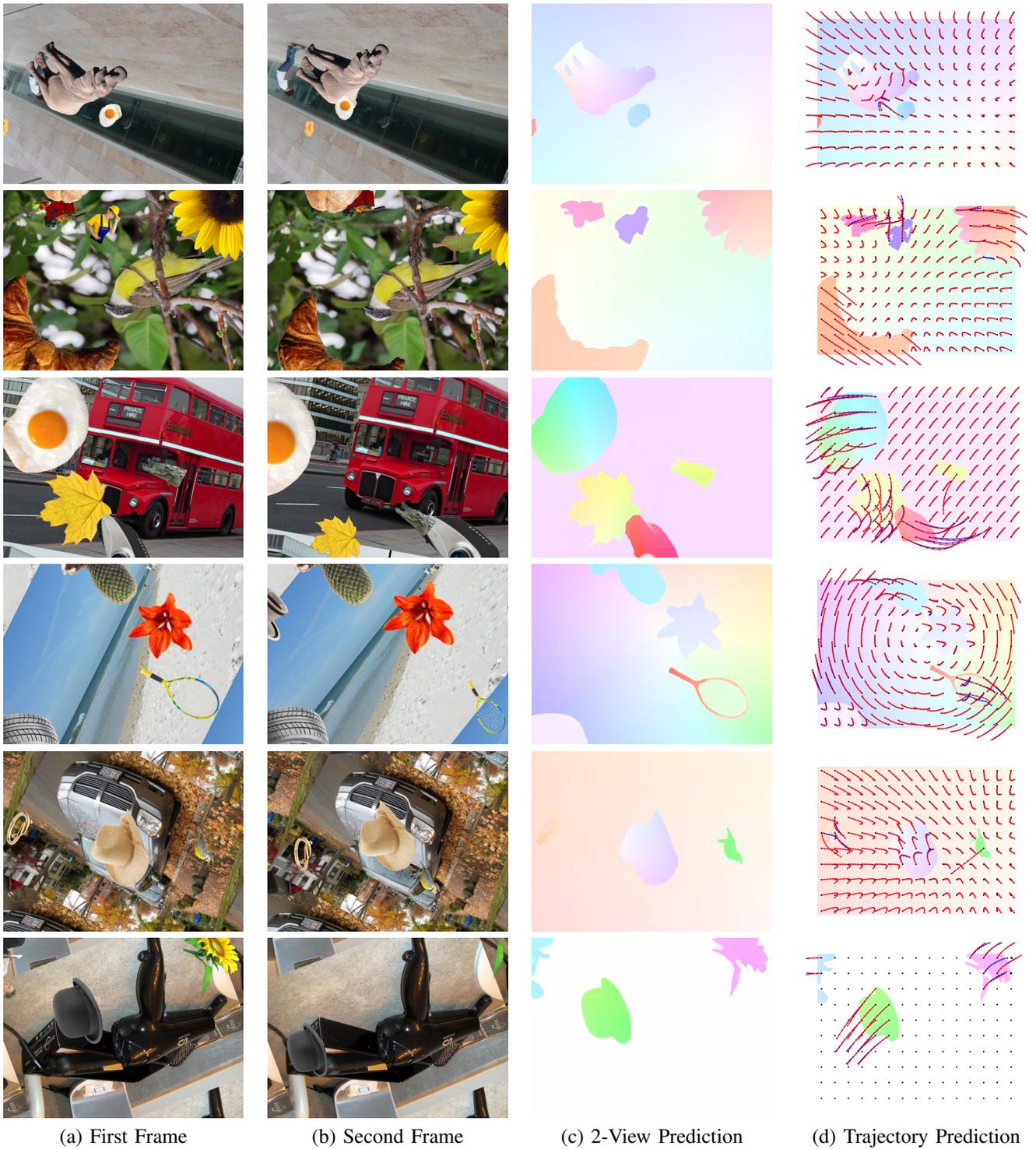

(a) First Frame     (b) Second Frame     (c) 2-View Prediction     (d) Trajectory Prediction

Fig. 3. Predictions of our model (E+I) on our MultiFlow test set. (a) shows the frame at the reference time, (b) the frame at the final timestamp of the trajectory, (c) shows the predicted pixel displacement between reference and final timestamp, and (d) shows the full prediction of the pixel trajectories. The predicted trajectory is shown in blue and the ground truth in red. The background of (d) is the colorization of the 2-view ground truth.

3



\end{document}

%% file: figs/eyecatch_page1.tex
\vspace{0.5cm}
\begin{tabular}{M{\imgfigWidthX}M{\imgfigWidthX}M{\thisfigWidthX}M{\imgfigWidthX}M{\imgfigWidthX}}
    \includegraphics[trim={4cm 25cm 69cm 20cm},clip,width=1.0\linewidth]{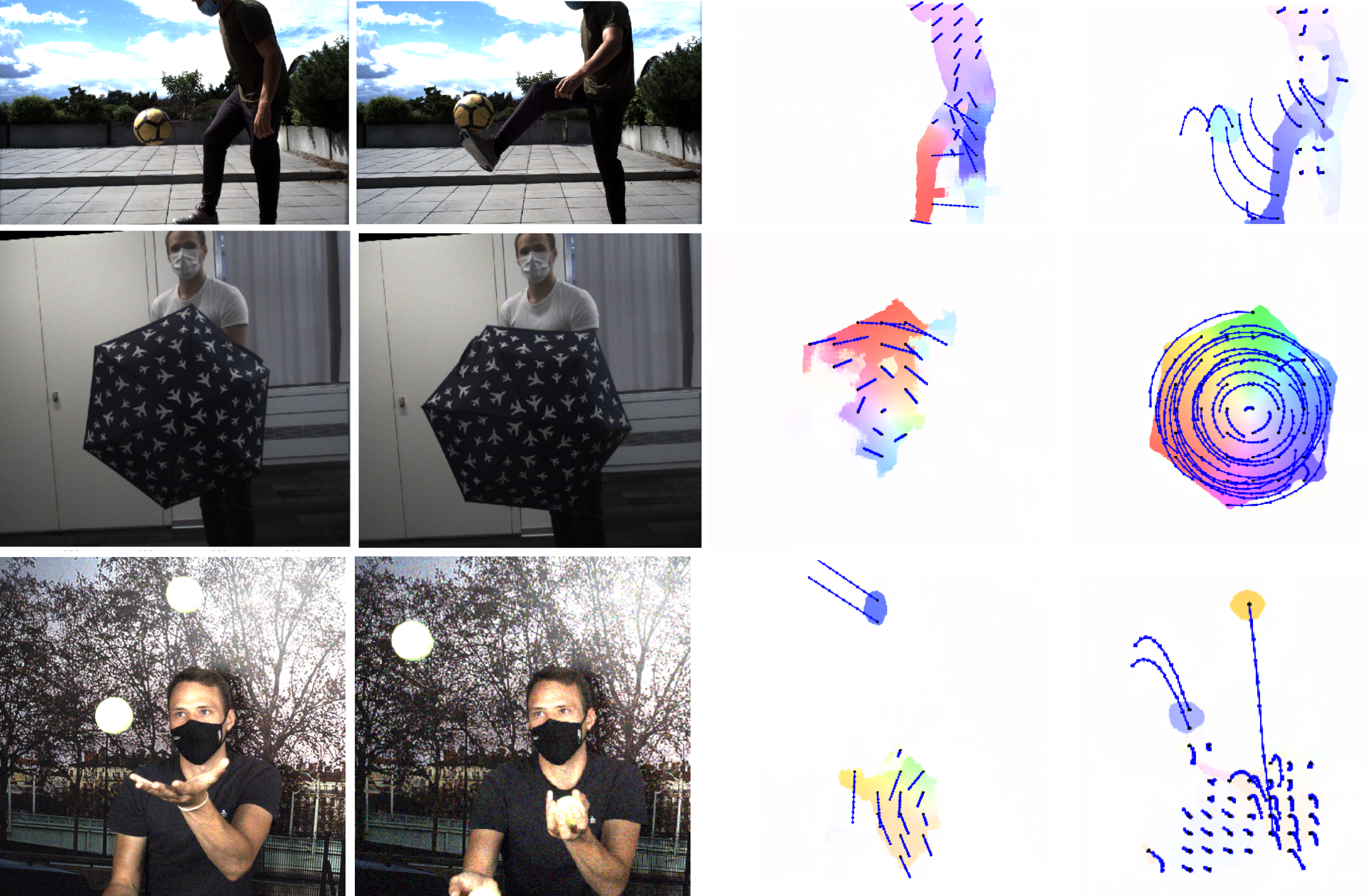} &
    \includegraphics[trim={26cm 25cm 47cm 20cm},clip,width=1.0\linewidth]{imgs/timelense/qualitative.png} &
    \includegraphics[trim={8.8cm 2cm 10cm 2cm},clip,width=1.0\linewidth]{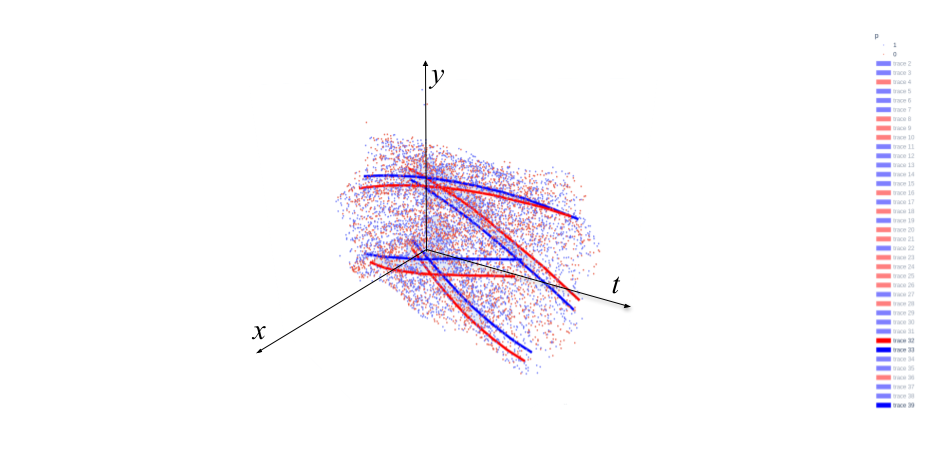}&
    \includegraphics[trim={72cm 25cm 1cm 20cm},clip,width=1.0\linewidth]{imgs/timelense/qualitative.png} &
    \includegraphics[trim={48cm 25cm 22cm 20cm},clip,width=1.0\linewidth]{imgs/timelense/qualitative.png}\\
    \footnotesize (a) First Frame & \footnotesize (b) Second Frame & \footnotesize (c) Events & \footnotesize (d)  Prediction & \footnotesize (e) Frame-based baseline\\
\end{tabular}
\captionof{figure}{(d) Our method, trained only on simulated events, predicts dense pixel trajectories in continuous-time on real sequences.
The frames in (a) and (b) are shown only for illustrative purposes. In (c), we also visualize selected pixel trajectories in the space-time volume of events. (e) The frame-based baseline \cite{Jiang_2021_ICCV} attempts to predict linear motion from the frames but falls short due to perceptual ambiguity and non-linear motion in pixel space.}
\label{fig:catcheye}

%% file: sections/abstract.tex
\begin{abstract}
    We present a method for estimating dense continuous-time optical flow from event data.
    Traditional dense optical flow methods compute the pixel displacement between two images.
    Due to missing information, these approaches cannot recover the pixel trajectories in the blind time between two images.
    In this work, we show that it is possible to compute per-pixel, continuous-time optical flow using events from an event camera.
    Events provide temporally fine-grained information about movement in pixel space due to their asynchronous nature and microsecond response time.
    We leverage these benefits to predict pixel trajectories densely in continuous time via parameterized \bez{} curves.
    To achieve this, we build a neural network with strong inductive biases for this task:
    First, we build multiple sequential correlation volumes in time using event data.
    Second, we use \bez{} curves to index these correlation volumes at multiple timestamps along the trajectory.
    Third, we use the retrieved correlation to update the \bez{} curve representations iteratively.
    Our method can optionally include image pairs to boost performance further.
    To the best of our knowledge, our model is the first method that can regress dense pixel trajectories from event data.
    To train and evaluate our model, we introduce a synthetic dataset (\simdset{}) that features moving objects and ground truth trajectories for every pixel.
    Our quantitative experiments not only suggest that our method successfully predicts pixel trajectories in continuous time but also that
    it is competitive in the traditional two-view pixel displacement metric on \simdset{} and DSEC-Flow. Open source code and datasets are released to the public.
\end{abstract}

%% file: sections/introduction_v2.tex
\section*{Multimedia Material}
\noindent \changes{Code and dataset available at:\\\url{https://github.com/uzh-rpg/bflow}}
\section{Introduction}\label{sec:introduction}

Optical flow estimation is a fundamental low-level vision task that informs about motion in pixel space.
It has numerous practical applications in computational photography and videography, video compression, inverse graphics, object tracking, and robotics \cite{guney2018optical}.

Traditionally, this problem has been addressed by finding dense correspondences between two frames.
Frame-based sensors provide information at a fixed frequency, independent of the dynamics in the scene, and must strike a trade-off between bandwidth and latency:
At high speeds, they require a high frame-rate to reduce perceptual latency, but this introduces a significant bandwidth-overhead for downstream systems.
Instead, reducing the frame-rate reduces the bandwidth requirements but at the cost of missing important scene dynamics.
These shortcomings have inspired recent work to address optical flow estimation with event cameras \cite{Zhu18rss,Gehrig3dv2021}.
In contrast to frame-based sensors, which capture frames at regular intervals, event cameras register per-pixel brightness changes asynchronously and at very high temporal resolution ($<1$ millisecond).
Additionally, they are robust to motion blur and have a very high dynamic range \cite{Gallego20pami}.
Due to these properties, event cameras are ideally suited for optical flow estimation.
Nonetheless, these advantages are tied to the fundamentally different data format resulting from event cameras.

Effectively extracting motion information from event cameras is a non-trivial task for which deep neural networks have shown promising results.
Until now, neural network approaches regress optical flow at a specific timestamp \cite{Zhu18rss,Gehrig3dv2021}.
Event cameras provide visual cues in continuous-time, which means that predicting discrete pixel displacements ignores much of the device's potential \cite{Gehrig3dv2021}.

In this work, instead, we go a step further and propose a differentiable model that can can exploit the rich spatio-temporal nature of event data by regressing continuous-time trajectories for every pixel of the camera.
Figure \ref{fig:catcheye} illustrates an example prediction of our method using only events in a short time window.
This approach not only enables new applications that can make use of continuous-time pixel trajectories \cite{Tulyakov21CVPR,Tulyakov_2022_CVPR} but also generally improves the accuracy of the model, as we will show in our experiments in Sec. \ref{sec:experiments}.

Extracting per-pixel continuous-time trajectories from event data is a challenging problem that requires careful modelling of the problem.
To achieve this, we propose the following methodological innovations:
\begin{itemize}
    \itemsep0em 
    \item Instead of predicting pixel displacements, we generalize this concept and estimate control points of \bez{} curves for each pixel.
    As a result, our method can regress the trajectory of each pixel at arbitrary times.
    It also enables the second innovation:
    \item We use multiple correlation volumes in time to search for pixel correspondences. We then use the estimated \bez{} curves to retrieve correlation features of all correlation volumes simultaneously. This enables the incorporation of motion prior and facilitates the task of finding accurate pixel trajectories.
    \item The use of image data is optional. We show that our approach works well purely with event data and enable dense, pixel-wise trajectories. Furthermore, we show that a combination of image and event data outperforms single-modality approaches on both real and synthetic datasets.
\end{itemize}
To train and evaluate our model we also contribute a new synthetic dataset \simdset{} which is inspired by the FlyingChairs dataset \cite{Dosovitskiy_2015_ICCV}.
Differently from the FlyingChairs dataset, \simdset{} features event data from various moving objects undergoing continuous similarity transformations as well as dense pixel-trajectory ground truth.

The proposed approach generalizes both the RAFT \cite{teed20eccv} and E-RAFT \cite{Gehrig3dv2021} architecture by not only giving the option to compute multiple correlation volumes in time but also to predict \bez{} curves that include the linear motion model as a special case.
The introduction of \bez{} curves gives the model the capabilities to estimate non-linear pixel-trajectories but also reduces the end-point-error that is used to assess the accuracy of predicted pixel displacements.

Our experiments suggest that simply giving the model the flexibility to predict \bez{} curves reduces the end-point-error on \simdset{} by approximately 67\%.
Additionally, the introduction of multiple correlation lookups in time further reduces the error metric computed on the whole trajectory by approximately 40\%.
Finally, we provide quantitative real-world experiments for the traditional optical flow task on DSEC-Flow \cite{Gehrig3dv2021}, reducing the end-point-error by 14.5\% compared to prior art.

%% file: sections/relwork.tex
\section{Related Work}

\subsection{Optical Flow}
For conciseness, we focus mostly on neural-network-based methods.

\subsubsection{Image-based}
The vast majority of neural network-based optical flow method considers the task of estimating dense pixel displacements from a pair of frames \cite{teed20eccv,Xu_2021_ICCV,Sun18cvpr,Ilg17cvpr,what_matters_uof,Wulff_2015_CVPR,Gadot_2016_CVPR,Bailer_2017_CVPR,Zhang21iccv,jaegle21arxiv,zhao2020maskflownet,Zheng_2020_CVPR,Dosovitskiy15iccv}.
A common component of many highly successful methods are explicit correlation volumes that guide the matching process.
This inductive bias enables high performance and data efficiency \cite{teed20eccv} as well as strong cross-dataset generalization \cite{Zhang21iccv}.

Multi-Frame optical flow estimation has been mostly explored in the self-supervised learning setting \cite{starflow20,Janai_2018_ECCV,proflow} or optimization-based literature \cite{Janai2017CVPR,ricco12cvpr,garg2013variational}.
Some neural-network-based approaches use an additional pair of frames to initialize the optical flow prediction \cite{teed20eccv} or use warped flow as input \cite{multi_frame_fusion_19} to a second stage.
Recent work proposes to track single pixels from video \cite{harley2022particle}.
By estimating the trajectory only for a single pixel, this method does not utilize spatial information but still achieves remarkable performance.
From these multi-frame approaches, Slow Flow \cite{Janai2017CVPR} is conceptually most related.
Slow Flow estimates optical flow $\mathbf{F}_{1\to N}$ given a sequence of $N$ images at high frame rate (1000 fps) with dense tracking using an optimization-based approach.
However, it requires a high-speed camera and complex occlusion handling that incurs a trade-off between drift and accurate prediction at motion boundaries.

\addtocounter{figure}{-1}
\input{figs/overview.tex}

\subsubsection{Event-based}
Event-based optical flow algorithms can be categorized into four categories: (i) Asynchronous methods \cite{Benosman12nn,Gehrig19ijcv} using the Lucas-Kanade algorithm \cite{Lucas81ijcai}.
(ii) Plane fitting-based methods that exploit the local plane-like shape of spatio-temporal event streams \cite{Butler:ECCV:2012,Mueggler15icra}.
(iii) Variational optimization-based approaches \cite{Bardow16cvpr,pan2020single} that incorporate image data\cite{pan2020single} or simultaneously  estimate image intensity \cite{Bardow16cvpr}.
(iv) Learning-based approaches, most of which are trained via self-supervision \cite{Zhu18rss,Zhu19cvpr,lee2020spike,paredes2020back,Ye21iros}. Supervision is provided either by images\cite{Zhu18rss,Zhu19cvpr,lee2020spike} or events\cite{paredes2020back,Ye21iros}.
Our approach is related to E-RAFT \cite{Gehrig3dv2021}, which adapts the RAFT \cite{teed20eccv} framework to event data to leverage correlation features from cost volumes to estimate dense pixel correspondences for large displacements.
In contrast to E-RAFT or RAFT, we predict pixel trajectories using multiple correlation lookups in time while we also show the advantages of combining events and frames.

\subsection{Continuous Tracking}

Continuous-time trajectory estimation of camera poses has been proposed in the context of rolling shutter compensation \cite{Kerl15iccv} as well as visual-intertial odometry for event cameras \cite{Mueggler18tro}.
Instead, we are interested in regressing pixel-trajectories, which is more closely related to high-speed feature tracking for event cameras \cite{Gehrig19ijcv,Seok2020,alzugaray2020haste,Zhu2017}.
Our approach is related to the work of Seok et al. \cite{Seok2020} who use quadratic \bez{} curves to sparsely track features by maximizing the variance of the image of warped events on local patches.
In contrast to our work, this method can only sparsely track features where events are present, cannot incorporate learned priors from data, and does not offer the possibility to include images.

\subsection{Datasets for Optical Flow}
Existing datasets can be categorized as image-based optical flow datasets and event-based optical flow datasets.

\subsubsection{Image-based} The seminal work of FlowNet \cite{Dosovitskiy15iccv} proposed a large synthetic dataset called "FlyingChairs" to train their CNN.
FlyingThings3D\cite{Mayer16cvpr} introduced a synthetic stereo-video dataset with scene flow ground truth.
AutoFlow\cite{Sun21cvpr} proposes to render large-scale training data with rich data augmentation in 2D for optical flow estimation.
These datasets, however, do not provide event data and ground truth pixel trajectories beyond two views.

\subsubsection{Event-based}
MVSEC \cite{Zhu18rss} contains 5 outdoor driving sequences and 4 indoor sequences.
However, optical flow ground truth from MVSEC suffers from inaccuracies in calibration \cite{Gehrig3dv2021} and only features very small displacements.
DSEC-Flow, a more recent dataset, addresses these shortcomings by providing accurate but sparse optical flow ground truth.
The main downside of MVSEC and DSEC-Flow is that they both do not include optical flow ground truth for dynamic objects.

In contrast to the aforementioned datasets, our synthetic dataset \emph{\simdset{}} features accurate and dense ground truth for pixel trajectories as well as image and event data. The concept of pixel trajectories is a generalization of pixel displacement and can be used to supervise and evaluate pixel tracking methods.

%% file: figs/overview.tex
\begin{figure*}[ht!]
    \centering
    \includegraphics[width=\linewidth]{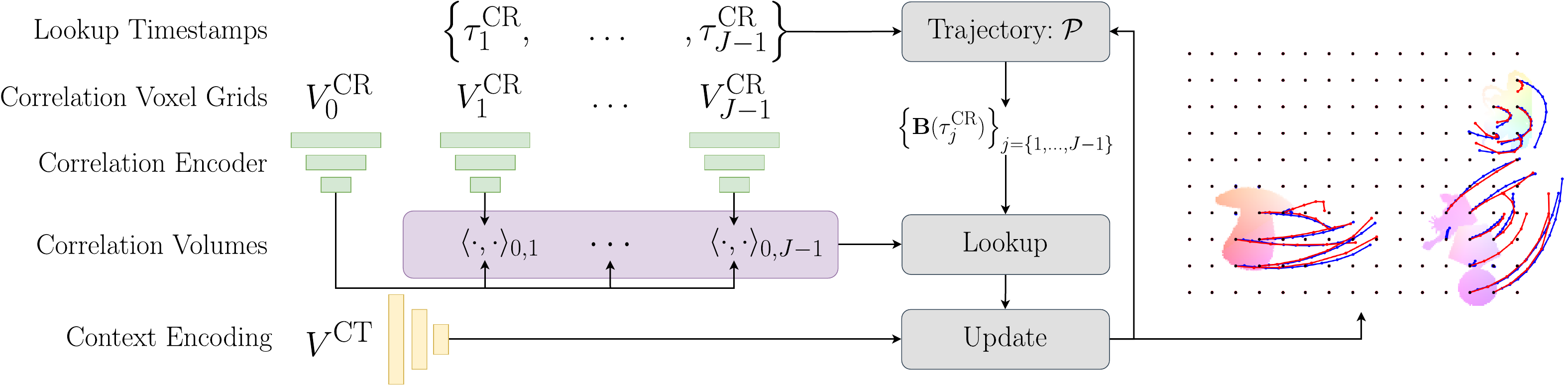}
    \caption{Overview of the framework. For concise presentation, we illustrate the purely event-based approach. We create multiple correlation voxel grids in a sequence to compute correlation volumes. \bez{} curves $\mathbf{B}$, parameterized by a set of control points $\mathcal{P}$, represent the continuous-time pixel trajectories. Along these trajectories, we index $J$ correlation volumes at their associated timestamps $\tau^\text{CR}$. An learned update operator uses the output of this lookup operation to update the \bez{} curves. Finally, this loop is repeated in an iterative fashion.}
    \label{fig:overview}
\end{figure*}

%% file: sections/methodology.tex
\section{Methodology}\label{sec:methodology}

\changes{
This paragraph outlines our methodology, setting the stage for the following detailed explanations. To assist in visualizing the process, Figure \ref{fig:overview} serves as a helpful guide. Our approach begins by transforming events into a spatio-temporal event representation, crucial for subsequent feature extraction, as elaborated in Section \ref{sec:input_data_prep}.
Following this, our model extracts two key types of features: context and correlation features from the event representations. The intricacies of this process are elaborated in Section \ref{sec:feat_ext}.
At the heart of our approach lies the computation of correlation volumes. These volumes are critical in assessing the quality of feature matches over time, thereby aiding in the tracking of pixel movements. This aspect is thoroughly explained in Section \ref{sec:multi_cost_volumes}.
Further, we describe our iterative approach to refine pixel motion, which is parameterized by Bézier curves initialized as straight lines. These curves are crucial for indexing the correlation volumes and thus, for guiding the trajectory estimation process. The nuances of this step are covered in Section \ref{sec:iterative_updates}.
}

\changes{
Our implementation reuses many components of the original RAFT architecture \cite{teed20eccv}.
While Figure \ref{fig:overview} offers a conceptual summary of our approach, the specific neural network architecture that integrates these elements is depicted in Figure 2 of the supplementary material.
}

\subsection{Problem Definition}

Our method is tasked with estimating a function
\begin{align}
    \mathbf{B}: \mathcal{T}\times\mathbb{N}_0\times\mathbb{N}_0 &\to \mathbb{R}^2\label{eq:cflow_def}\\
    (\tau, x, y) &\mapsto \mathbf{B}(\tau, x, y)
\end{align}
that describes the per-pixel trajectories in time on the image plane.
$\mathcal{T}=\mathbb{R}\cap[0,1]$ describes the domain of the normalized time $\tau(t) = \nicefrac{(t - t_\rf)}{(t_\tg - t_\rf)}$
, where $\tau = 0$ corresponds to the reference time $t_\rf$ from which the pixel trajectory starts and $\tau=1$ corresponds to the target time $t_\tg$ when the pixel trajectory ends.
This formulation can be used to find the pixel displacement at any time between $t_\rf$ and $t_\tg$.

\changes{
As will be explained in more detail later, the proposed approach also uses additional events from time $t_0$ until the reference time $t_\rf$ to extract feature maps at different timestamps.
However, the pixel trajectory is estimated only from the reference time $t_\rf$ to the target time $t_\tg$.
}

\subsection{Input Data Preparation}\label{sec:input_data_prep}
The proposed approach uses features extracted from event data and optionally a pair of images to further boost performance.

\subsubsection{Multi-View Event Representations}\label{sec:multi_view_ev_repr}
Event cameras have a fundamentally different working principle than frame-based cameras.
Instead of acquiring frames, event cameras receive asynchronous events with high temporal resolution.
An event $e_k(t) = (x_k, y_k, t, p_k)$ is a tuple containing information about a pixel $(x_k, y_k)$ for which a positive or negative brightness change $p_k$ was registered at time $t$. Note that the time $t$ typically has microsecond resolution, which provides precise temporal information about motion in the scene.

\input{figs/ev_repr_gen.tex}

The first step of the feature extraction pipeline is the construction of a discrete spatio-temporal representation from a sequence of events.
For our experiments, we choose the voxel grid representation by Zhu et al. \cite{Zhu19cvpr} due to its simplicity and possibility to extract features in a sliding window.
Figure \ref{fig:voxel_gen} provides an overiew of this process.

Given the task of estimating continuous pixel trajectories from $t_\rf$, the reference time, to the target time $t_\tg$,
Our method first computes a \emph{base} voxel grid, visualized in red in Figure \ref{fig:voxel_gen}, consisting of $M+N-1$ discrete bins along the time dimension via interpolation of event data \cite{Zhu19cvpr}.
The first $M$ bins are computed from events between initial time $t_0$ and the reference time $t_\rf$. These events are used to extract the correlation features later in the pipeline.
The last $N$ bins are computed from events during the time window $[t_\tg, t_\rf]$ when the pixel trajectories are estimated.
These $N$ bins are required to extract the context features.
The bin at the reference timestamp $t_\rf$ will be reused by both correlation and context features such that the base voxel grid consists of $M + N - 1$ bins.

From this base voxel grid, we first extract the context voxel grid $V^\cnt$, visualized in yellow in Figure \ref{fig:voxel_gen}.
The next step is the extraction of $J \leq N$ correlation voxel grids $V^\crr_j$, visualized in green in Figure \ref{fig:voxel_gen}, in a sliding-window fashion from the base voxel grid.
While it is possible to extract up to $J=N$ correlation voxel grids, we find in our ablation study in Section \ref{sec:exp:num_coor_lookups} that the choice of $J$ can be used to trade off performance and compute.
Each of these correlation voxel grids contain information about the contrast differences in the scene at slightly different times.
They will later be used to guide the search for pixel trajectories via lookup operations.
We refer to the correlation voxel grids also as \emph{views} because they each represent a distinct timestamp.
If we choose to extract at least two correlation voxel grids apart from the reference timestamp $t_\rf$ (i.e. $J\geq 3$), we categorize the method as \emph{multi-view}.

\subsubsection{Optional Frame-based Input}
As we show in the experimental results, the performance of the proposed method improves if image data is used in addition to events only.
To do so, we acquire a reference frame $I_\rf$ at $t_\rf$ and a target frame $I_\tg$ at $t_\tg$.
These frames contain richer information about texture at the boundary timestamps of the regressed pixel trajectories and thus simplify the correspondence search.

\subsection{Feature Extraction}\label{sec:feat_ext}
Features are extracted from input images and voxel grids with a convolutional network.
The basic architecture of the encoders follows prior work \cite{teed20eccv,Gehrig3dv2021}.
The encoders extract $D=256$ dimensional features from their input at $1/8$th of the original resolution using residual blocks with striding for downsampling the feature maps.
From now on, we use $H'= H/8$ and $W'=W/8$ to refer to the downsampled resolution.

The context feature encoder $f_\cnt: \mathbb{R}^{(N+3)\times H\times W}\mapsto \mathbb{R}^{D\times H'\times W'}$ concatenates the voxel grid $V^\cnt$ with the frame $I_\rf$, if available, to extract combined features.
The reference frame $I_\rf$ informs the network about the absolute intensity of the reference pixels while the context voxel grid $V^\cnt$ provides rich information about motion during the time duration of the pixel trajectories.
\changes{Note that the context network only extracts features from  $I_\rf$ and not $I_\tg$ because we require the context features to be aligned with the reference timestamp.}
Finally, contrast information is not explicitly considered yet, which will be provided by the correlation feature encoders.

The correlation feature encoder $f_\crr^V: \mathbb{R}^{M\times H\times W}\mapsto \mathbb{R}^{D\times H'\times W'}$ for event representations computes features from the $N$ correlation voxel grids in parallel, by sharing the weights.
The result of this operation are $N$ feature maps, each assigned to the timestamp associated with the last bin of each correlation voxel grid.
If in addition, a pair of frames is available, we extract image features with an additional encoder $f_\crr^I: \mathbb{R}^{3\times H\times W}\mapsto \mathbb{R}^{D\times H'\times W'}$.

\subsection{Multi-View Correlation Volumes}\label{sec:multi_cost_volumes}
We compute correlation volumes to guide correspondence search in space and time by associating subsequent views with the reference view.
This association is visualized in Figure \ref{fig:corr_lookup} and \ref{fig:overview}.
We use the features created from voxel grid $V_0^{CR}$ and optionally $I_\text{r}$ to create the feature maps for the reference view.
For each subsequent view $j$, we compute features from voxel grid $V_j^{CR}$ to compute the correlation volume as in equations \eqref{eq:corr_compute} and \eqref{eq:corr_compute_v}.
For the final/target view, we optionally compute a correlation volume from the boundary image features as in equations \eqref{eq:corr_compute} and \eqref{eq:corr_compute_i}.
We perform the computation of all correlation volumes in parallel with batched matrix multiplication.

\begin{align}
    \corrv\left(f_\crr^V(V_0^\crr), f_\crr^V(V_j^\crr)\right) &\in\mathbb{R}^{H'\times W'\times H'\times W'}\label{eq:corr_compute_v},\\
    \corrv\left(f_\crr^I (I_\rf), f_\crr^I (I_\tg)\right) &\in\mathbb{R}^{H'\times W'\times H'\times W'}\label{eq:corr_compute_i},
\end{align}
where
\begin{align}
    C_{ijkl}(a, b) &= \frac{1}{\sqrt{D}}\sum_d a_{dij} \cdot b_{dkl}\label{eq:corr_compute}
\end{align}

\subsection{Iterative Multi-View Flow Updates}\label{sec:iterative_updates}
This section describes the update scheme that uses correlation volumes introduced in the previous section \ref{sec:multi_cost_volumes}.

\input{figs/corr_lookup.tex}

\subsubsection{Continuous-Time Flow}\label{sec:cont_flow}
We seek a representation of the function $\mathbf{B}(\tau, x, y)$, as introduced in equation \eqref{eq:cflow_def}, that generalizes the conventional displacement prediction of two-frame optical flow methods.
To achieve this, we choose to use \bez{} curves defined as
\begin{equation}
    \mathbf{B}(\tau, x, y) = \sum_{i=0}^n\begin{pmatrix}n\\i\end{pmatrix}(1-\tau)^{n-i}\tau^i\mathbf{P}_i(x, y).\; 
\end{equation}
$\mathbf{B}(\tau, x, y)$ describes the displacement of pixel $(x, y)$ for the normalized time $\tau\in[0, 1]$.
As an example, the \bez{} curves of degree $n=1$  simply represent a linear trajectories in space and time that is fully described by $\mathbf{P}_1$. The task of our method, however, is to regress the parameter set
\begin{equation}
    \mathcal{P} = \left\{\mathbf{P}_1, \ldots, \mathbf{P}_n\right\}
\end{equation}
$\mathbf{P}_0 = \mathbf{0}$ does not have to be estimated because it defines the starting point of the trajectory that coincides with the reference pixels.
The advantages of working with \bez{} curves is that they are fast to evaluate and can be concisely summarized by a set of parameters $\mathcal{P}$.

The degree $n$ of the \bez{} curves is a fixed parameter.
Our ablation study in Section \ref{sec:exp:bez_degree} suggests to set $n$ equal to the number of supervision points along the trajectory for an accurate trajectory prediction.

\textbf{Iterative \bez{} Updates:}
The \bez{} curves are initialized with $\mathcal{P}=\{\mathbf{0}, \ldots, \mathbf{0}\}$ such that $\mathbf{B}=\mathbf{0}$.
The update block of the network estimates increments $\Delta\mathbf{P}$ that are added to the current estimate of the \bez{} curve parameters.
More specifically, in each iteration from $k$ to $k+1$ the network updates the parameter set $\mathcal{P}$ the following way:
\begin{align*}
    \mathbf{P}_i^{k+1} = \mathbf{P}_i^{k} + \Delta\mathbf{P}_i^k,\;\forall i\in\{1, \ldots, n\}
\end{align*}

\subsubsection{Multi-View Correlation Lookup}
Similar to the original RAFT implementation, we use lookup operations to extract features from the correlation volumes.
In contrast, however, we extract features from multiple correlation volumes, each associated to a unique normalized timestamp.
Given the current estimate of the \bez{} control points $\mathcal{P}$, we map each pixel $\mathbf{x}=(x, y)$ at time $\tau=0$ to the estimated corresponding pixel location $\mathbf{x}'(\tau)=(x'(\tau), y'(\tau))$ at time $\tau$:
\begin{equation}
    \mathbf{x}'(\tau) = \mathbf{x} + \mathbf{B}(\tau, \mathbf{x})
\end{equation}
Similar to RAFT, the lookup is performed in a local neighborhood $\mathcal{N}$ around the corresponding pixel location $\mathbf{x}'(\tau)$:
\begin{equation}
    \mathcal{N}(\mathbf{x}'(\tau)) = \left\{\mathbf{x}'(\tau) + \mathbf{dx}\,\vert\, \mathbf{dx}\in\mathbb{Z}^2, ||\mathbf{dx}||_\infty \leq r\right\}
\end{equation}
Lookups are performed on all available correlation volumes with bilinear sampling. We use a constant lookup radius of $r=4$, as in the original RAFT implementation, to increase the effective lookup radius.
Finally, the values from the union of lookup operations are concatenated into a single feature map.

\subsubsection{Upsampling of \bez{} Curves}
The \bez{} control points are estimated at $\nicefrac{1}{8}$-th of the original resolution. Since $\mathbf{B}(\cdot)$ is linear in the control points $\mathcal{P}$, we can upsample the \bez{} curves to the full resolution using convex upsampling \cite{teed20eccv}. As a result, the \bez{} curves at the full resolution will be a learned convex combination of $3\times 3$ grids of \bez{} curves at the lower resolution.

\subsection{Supervision}
\input{figs/loss_fun.tex}
We supervise the model with $N_k$ ground truth flow maps along the trajectory, visualized in Figure \ref{fig:loss_function}.

\begin{equation}
    \mathcal{L} = \frac{1}{N_k}\sum_{i=1}^{N_i}\gamma^{N_i-i}\sum_{k=1}^{N_k}||\mathbf{f}_{gt}(\tau_k) - \mathbf{B}_i(\tau_k)||_1\label{eq:loss}
\end{equation}
where $\mathbf{B}_i$ is the \bez{} curve at iteration $i$, $\tau_k\in[0, 1]$ are the evaluation timestamps of the \bez{} curves, and $\gamma=0.8$.
For a single displacement map, such as in two-frame optical flow, the corresponding parameters are $N_k=1$ and $\tau_1=1$.

%% file: figs/ev_repr_gen.tex
\begin{figure}[t!]
    \centering
    \includegraphics[width=1.00\linewidth]{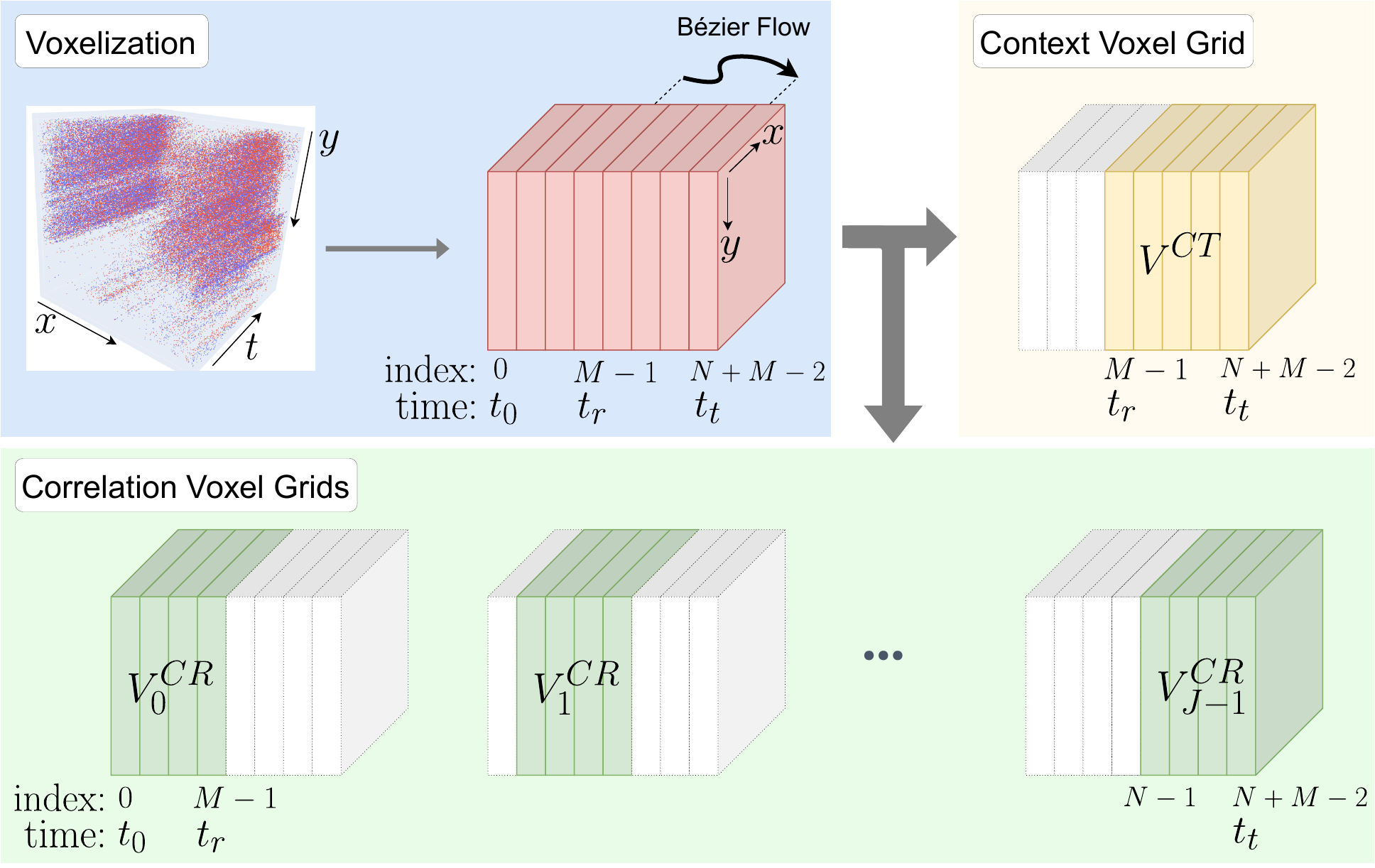}
    \caption{Voxel Grid Generation. Events are interpolated into a voxel grid representation before being divided into sub-voxel grids that are either used as input to the context encoder or the correlation encoder. The \bez{} curves, indicated in the figure, are estimated from the reference time $t_r$ to the target time $t_t$ for each pixel.
    The initial voxel grid channels before the reference time $t_r$ are used for extracting correlation voxel grids as visualized in the lowest, green row of the figure.}
    \label{fig:voxel_gen}
\end{figure}

%% file: figs/corr_lookup.tex
\begin{figure}[t]
    \centering
    \includegraphics[width=0.6\linewidth]{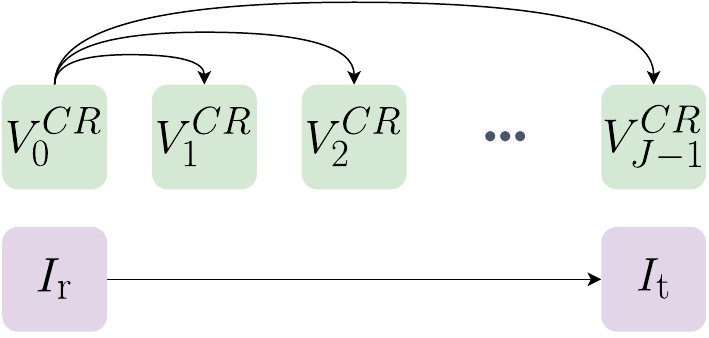}
    \caption{Correlation-volume pairs with indicated lookup direction. The upper row shows lookups from voxel grid features and the lower row shows lookups from image features.}
    \label{fig:corr_lookup}
\end{figure}

%% file: figs/loss_fun.tex
\begin{figure}[t]
    \centering
    \includegraphics[width=1.00\linewidth]{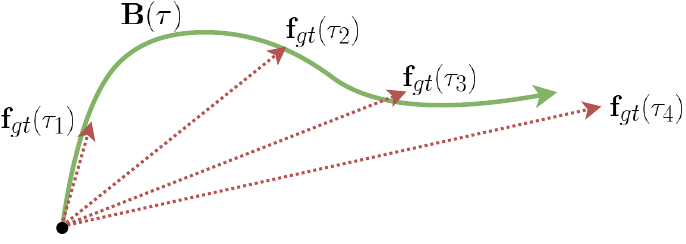}
    \caption{Illustrative example of a \bez{} prediction and $N_k=4$ ground truth flow maps from which the loss function (\ref{eq:loss}) is computed.}
    \label{fig:loss_function}
\end{figure}

%% file: sections/dataset.tex
\section{\simdset{} Dataset}
\label{sec:dataset}
To the best of our knowledge, there are no publicly available datasets with dense ground truth for pixel trajectories, which is more general than pixel displacements, in combination with events and images.
Therefore, we create a new synthetic dataset to evaluate the proposed approach.

\subsection{Data Generation Procedure}
We generate 10100 training and 2000 test sequences in simulation for \simdset.
The background images are sampled from the Flickr30K dataset \cite{young2014image}.
The foreground objects are extracted from randomly sampled PNG images, by masking out transparent regions using their alpha channels.

The duration of each sequence is defined to be one second.
We randomly sample 3 or 4 control points over similarity transformations in 2D for each foreground and background object.
Interpolation with piece-wise cubic spline polynomials provide the continuous-time transformation of all objects throughout the duration of each sequence.
The generated similarity transformation trajectories are used to both render images and compute ground truth pixel trajectories.
To generate events, we apply the generative model of events \cite{Rebecq18corl} on frames rendered at 1000 frames per second on the full sequence.

\textbf{Pixel Trajectory Ground Truth: }
The ground truth for the pixel trajectories is computed with respect to a reference time at $0.4$ seconds.
The pixel trajectory is expressed as pixel displacements at different target timestamps with respect to the original pixel location at the reference time.
Overall, the dataset consists of pixel trajectory ground truth at intervals of 10 milliseconds from $0.4$ seconds ($t_\rf$) up to time $0.9$ seconds ($t_\tg$).
Events outside the time interval where ground truth is available ($0.4\leq t \leq 0.9$) can be used as additional context.
We refer to the supplementary material for dataset examples and a more detailed description.

%% file: sections/experiments.tex
\input{figs/multiflow_examples.tex}
\section{Experiments}\label{sec:experiments}

This section has four distinct purposes.
First, to show quantitative real-world performance on DSEC-Flow on traditional two-view metrics.
Second, to show quantitative performance on the \emph{main task} on \simdset{}: predicting continuous-time pixel trajectories.
Third, to show qualitative performance for sim2real transfer using models trained on our \simdset{} dataset.
Finally, an ablation study highlights the importance of the proposed components.

\textbf{Implementation Details: }
Our models are implemented in Pytorch and trained from scratch with random weights on each dataset. We use AdamW \cite{loshchilov2018decoupled} with gradient clipping in the range of $[-\mathbf{1}, \mathbf{1}]$ and a OneCycle learning rate with a batch size of 3. On both datasets, we perform random horizontal and vertical flipping as well as random cropping of the input data.

\input{tables/dsec_flow_exp}
\subsection{DSEC-Flow}
DSEC-Flow \cite{Gehrig3dv2021,gehrig2021dsec} is a driving dataset with stereo event and global shutter cameras.
The purpose of this experiment is to show the quantitative performance of our approach on real-world data. However, we can only test on traditional two-view optical flow metrics because no pixel trajectory ground truth is available on DSEC-Flow.
In other words, for displacement prediction, the quality of the intermediate continuous trajectory is irrelevant.
Instead, only the final displacement prediction is evaluated using end-point-error (EPE) and angular error (AE) metrics \cite{Baker11ijcv}.
We also report the $X$-point error, a metric that reports the percentage of pixels with EPE higher than $X$ pixels.

For our experiments that combine image and event data, we warp images to the event camera.
Due to the small baseline between both cameras, we can warp the image to a plane at an infinite depth and re-project it into the event camera coordinate frame.
The disparity is negligible.

The performance of EV-FlowNet \cite{Zhu18rss} and E-RAFT is taken from Gehrig et al. \cite{Gehrig3dv2021}. We additionally train the frame-based RAFT model \cite{teed20eccv} and GMA \cite{Jiang_2021_ICCV}, an addition over RAFT, to also compare against purely frame-based approaches.

\textbf{Implementation Details: }
Our model is supervised with the 2-view version of our loss proposed in equation \eqref{eq:loss}.
This loss function is used because DSEC-Flow only provides ground truth for pixel displacements.
We use $M=N=5$ bins for both the context and correlation voxel grid (see Section \ref{sec:multi_view_ev_repr}) and $J=5$ views to compute the correlation volumes. %
We also experimented with a higher number of bins but did not observe significant improvements.
Finally, we choose a degree of 2 for the \bez{} curves mostly to account for non-linear motion due to change in depth.
We train our models for 250k iterations on a single Titan RTX which takes up to 40 hours.

\subsubsection{Quantitative Evaluation}
Table \ref{tab:dsec_flow_exp} summarizes our results on the DSEC-Flow test set.
Our approach, using both event data and frames, achieves 0.69 EPE which is 11.5\% lower than RAFT \cite{teed20eccv} and 8 \% lower than our own method using only event data.
Furthermore, our purely event-based approach achieves an EPE of 0.75 which is 5\% lower than the EPE of 0.79 that E-RAFT achieves.
Overall the performance of E-RAFT \cite{Gehrig3dv2021} is comparable to RAFT while EV-FlowNet \cite{Zhu18rss} is not competitive.
In our experiments, the GMA version of RAFT does not outperform the RAFT baseline on this dataset.

These results indicate that the proposed method, even though designed to work for regressing pixel trajectories, is competitive with two-view approaches on the task of pixel displacement prediction.
Note that we have not used any additional ground truth information compared to the baselines.

\input{tables/multiflow_exp}

\subsection{\simdset}\label{sec:multiflow_exp}
The experiments on \simdset{} assess the pixel trajectory regression capabilities.
To achieve this, we introduce an extension of EPE and AE to trajectories.
\begin{align}
    \text{TEPE} &= \frac{1}{N_k}\sum_k^{N_k}\text{EPE}(\mathbf{f}_{pred}(t_k),\mathbf{f}_{gt}(t_k)),
\end{align}
where $N_k \geq 2$. We analogously define TAE..%

We train three previously published baselines for an extensive comparison.
First, RAFT \cite{teed20eccv} and RAFT+GMA \cite{Jiang_2021_ICCV} for a comparison against frame-based approaches.
Second, E-RAFT \cite{Gehrig3dv2021} for a comparison against a recent event-based approach.
We focus on these architectures because they are related to our approach and achieve competitive performance on public benchmarks \cite{Butler12eccv,gehrig2021dsec}.

\textbf{Implementation Details: }
We train our method on the loss defined by equation \eqref{eq:loss} while the two-view approaches are trained on the two-view version of the loss \cite{teed20eccv}\footnote{It is also possible to use equation \eqref{eq:loss} to supervise two-view methods, but it results in an unfavorable trade-off between trajectory end-point-error error (TEPE) and end-point-error (EPE).}.
We supervise our methods with $10$ flow maps along the trajectory using loss \eqref{eq:loss} and set the \bez{} curve degree to 10 according to section \ref{sec:cont_flow}.
The ablation study in Section \ref{sec:exp:bez_degree} clarifies the relationship between the loss function \eqref{eq:loss} and the degree of the \bez{} curve.

For the following experiments, we build $J=6$ correlation voxel grids representing the views at regular time intervals starting from 0.4 seconds up to 0.9 seconds (i.e. the duration where we have to predict the pixel trajectories in the dataset).
We discretize the time into regular bins resulting in $N=41$ bins for the context voxel grid and $M=25$ bins for the correlation voxel grids.
Later in section \ref{sec:voxel_bin_size}, we show that a coarse discretization with fewer bins leads to suboptimal performance.
We observed that the reason for this is the overwriting of polarities in the voxel grid which we prevent by choosing a more fine-grained temporal discretization.

We train our models for 200k iterations on a single Titan RTX which takes up to 80 hours.

\input{figs/timelens_examples_v2.tex}

\subsubsection{Quantitative Evaluation}
\changes{
In addition to E-RAFT and RAFT variants using frames, we also compare against a baseline that consists of the E-RAFT approach that estimates \bez{} curves instead of pixel displacements, as in the original work.
This approach, named \emph{E-RAFT + \bez{}}, is better suited for non-linear flow estimation than prior work.
Table \ref{tab:multiflow_exp} summarizes the quantitative results.
Trajectory metrics in brackets indicate approaches that predict pixel displacements (RAFT, RAFT + GMA, and E-RAFT), for which we employ linear interpolation to compute the trajectory metrics.
As expected, the results suggest that these approaches are unsuitable for accurate pixel trajectory estimation.
The \emph{E-RAFT + \bez{}} baseline clearly improves not only the trajectory metrics but also the 2-view metrics compared to these baselines.
Our proposed model substantially reduces the TEPE further from 2.62 to 1.85 by using correlation features to guide the trajectory estimation.
Finally, our approach benefits from combining events and frames: TEPE is reduced from 1.85 to 1.29 by 30\% while the EPE also decreases by 32\%.
}

\changes{
When purely comparing 2-view metrics, RAFT + GMA \cite{Jiang_2021_ICCV} achieves the lowest error by extending RAFT with a self-attention mechanism on the context features to aggregate motion features.
However, RAFT+GMA is not designed for trajectory prediction and, therefore, falls short in TEPE and TAE compared to our approach.
}

\subsubsection{Qualitative Analysis}
Figure \ref{fig:multiflow_examples} illustrates the predictions of our model, using events and frames, in comparison to RAFT+GMA \cite{Jiang_2021_ICCV}.
Our approaches successfully predict the continuous trajectory, even if the objects leave the field of view.
This is not the case for the two-view method RAFT+GMA, as seen in the first row of Figure \ref{fig:multiflow_examples}.
Evidently, it fails at predicting the motion of objects that are missing in the second view because it is impossible to establish a match. 
The second row shows that our method can accurately predict pixel trajectories while RAFT+GMA does well in predicting the final pixel displacement.

\subsection{Qualitative Sim2Real Results}
To qualitatively assess the real-world capabilities of our method, we use our purely event-based model, trained only on the simulation dataset \simdset{}, and show predictions on the \tldata{} and BS-ERGB datasets \cite{Tulyakov21CVPR,Tulyakov_2022_CVPR} in Figure \ref{fig:catcheye} and \ref{fig:timelense_examples}.
Although our method is only trained on \simdset{}, it can correctly predict non-linear motion of moving objects while the frame-based baseline \cite{Jiang_2021_ICCV} fails due to large motion and ambiguities in the input frames.

\input{tables/ablation_exp}

\subsection{Ablation Study}
This section examines the influence of the main contributions of our model as well as additional design choices that contribute to the performance of the method.
We perform the ablation studies on the \simdset{} because we require accurate pixel trajectory ground truth that is not available on DSEC or other event-based datasets.

\subsubsection{Main Components}
One of the main contributions of this paper are the iterative \bez{} curve regression and the correlation lookup applied at multiple timesteps/views along the trajectories.
We ablate both components using the full model described in Section \ref{sec:methodology} as reference and summarize the results in Table \ref{tab:ablation_exp}.

First, we remove the \bez{} curve regression and note a drastic drop in performance on all metrics.
Removing the \bez{} curves is equivalent to assuming linear motion.
Linear motion is not representative for this dataset and impairs the utility of the correlation lookup.
For example, if the model is constrained to predict a linear trajectory during non-linear motion, the correlation lookup will not be accurately placed along the trajectory. This leads to correlation features that are not informative.

Second, we remove the correlation lookups between the reference view and intermediate views, denoted as the w/o multi-view baseline in Table \ref{tab:ablation_exp}, and note that both TEPE and EPE increase between 18\% and 42\% with respect to the event-based and hybrid reference model.
We conclude that both components are essential for achieving the best results.

\subsubsection{Degree of the \bez{} Curves}\label{sec:exp:bez_degree}
The goal of this experiment is to answer the question: \emph{Which \bez{} curve degree is appropriate for our model?}

To answer this question, we train 5 different models with \bez{} curve degrees 1, 5, 10, 15, and 20 using frames and events as input to the model. The supervision on pixel trajectories at training time is at 20 Hz, that we supervise the model on the pixel trajectory every 50 milliseconds.

\input{figs/bez_degree.tex}

Intuitively, one could surmise that a higher \bez{} degree always improves the performance. Figure \ref{fig:bez_degree} indeed shows that a higher \bez{} degree improves performance, when the models are evaluated at 20 Hz.
The models were also trained with a supervision signal at 20 Hz, that is with supervision at regular time intervals of 50 milliseconds.
Interestingly, an evaluation of these models at a higher frequency of 100 Hz reveals a decrease in performance for the model with a \bez{} degree higher than 10.
The reason for that is that we observe a test-time overfitting of the predicted trajectory to the frequency with which the model was supervised.
We do not observe this overfitting issue at 20 Hz evaluation frequency because the model is able to accurately predict the trajectory at this frequency.
This observation is reminiscent of the overfitting phenomenon in the much simpler polynomial regression problem in statistics.

The answer to the original question is: \emph{For accurate pixel trajectory predictions, the \bez{} degree should be chosen equal to the number of regular supervision points along the pixel trajectories.}
The degree can be set higher than the number of supervision points to potentially improve performance on the timestamps of interest.
However, the pixel trajectories may not be as accurate anymore.
An example for this is our DSEC-Flow experiments where we have only 1 supervision point (2-view) but found that a \bez{} degree of 2 improves the performance on the 2-view metric.
This is a reasonable approach as long as the downstream application does not require accurate pixel trajectories.

\subsubsection{Loss Function}
This experiment examines the influence of the trajectory loss on trajectory and 2-view metrics.
We train two additional models with the 2-view loss; one using only events and a second one using both frames and events.
Table \ref{tab:ablation_loss} summarizes the results.

For both input variations, the error metrics substantially decrease when the model is trained with the trajectory loss.
This reduction is evident for both the trajectory metrics and also the 2-view metrics.
This indicates, that the trajectory loss function is a better choice for the proposed method than the 2-view loss.
Finally, the results in Table \ref{tab:ablation_loss} indicate that we could further reduce the errors of the DSEC-Flow experiment with the appropriate ground truth.

\input{tables/supp/ablation_loss.tex}

\subsubsection{Number of Correlation Lookups}\label{sec:exp:num_coor_lookups}
Table \ref{tab:ablation_num_corr} shows that increasing the number of correlation lookups improves performance consistently on all metrics.
However, substantially increasing the number of correlation volumes incurs higher memory consumption and increases computational demand.
This result suggests a trade-off between performance and compute and memory requirements.
Note that the number of correlation volumes is the number of correlation lookups (in time) + 1 because the correlation volume at the reference time has to be taken into account as well.

\input{tables/supp/ablation_num_corr.tex}

\changes{
\subsubsection{Voxel Grid Temporal Resolution}\label{sec:voxel_bin_size}
The voxel grid serves as a discrete representation of event data, 
inevitably leading to some loss of information during its formation.
In particular, a decrease in the number of temporal bins not only reduces the number of input channels but also intensifies the quantization of the signal.
In our study, detailed in Table \ref{tab:ablation_voxel_bins}, we investigate how the temporal resolution of the voxel grid affects performance on the MultiFlow dataset.
The bin size in this context refers to the temporal interval between consecutive bins.
Consistent with expectations, Table \ref{tab:ablation_voxel_bins} demonstrates that a finer bin size, corresponding to higher temporal resolution, results in improved performance.
Note that a bin size of 0.0125 represents the voxel grid resolution, as detailed in section \ref{sec:multiflow_exp}, which was used for the main results on the MultiFlow dataset.
}

\input{tables/supp/ablation_voxel_bins}

\subsection{Runtime Analysis}
Table \ref{tab:model_stats} shows inference time in milliseconds, parameter count in millions of parameters and memory consumption at inference time on samples from the \simdset{} dataset.
Overall, our proposed method has higher demands on the presented metrics compared to RAFT \cite{teed20eccv}, which is expected because it is a generalization of the RAFT architecture.

\input{tables/supp/model_stats.tex}

%% file: figs/multiflow_examples.tex
\begin{figure*}[!t]
    \centering
    \newcommand{\thisfigWidth}{0.230\linewidth}
    \begin{tabular}{M{\thisfigWidth}M{\thisfigWidth}M{\thisfigWidth}M{\thisfigWidth}}
        \includegraphics[width=0.84\linewidth]{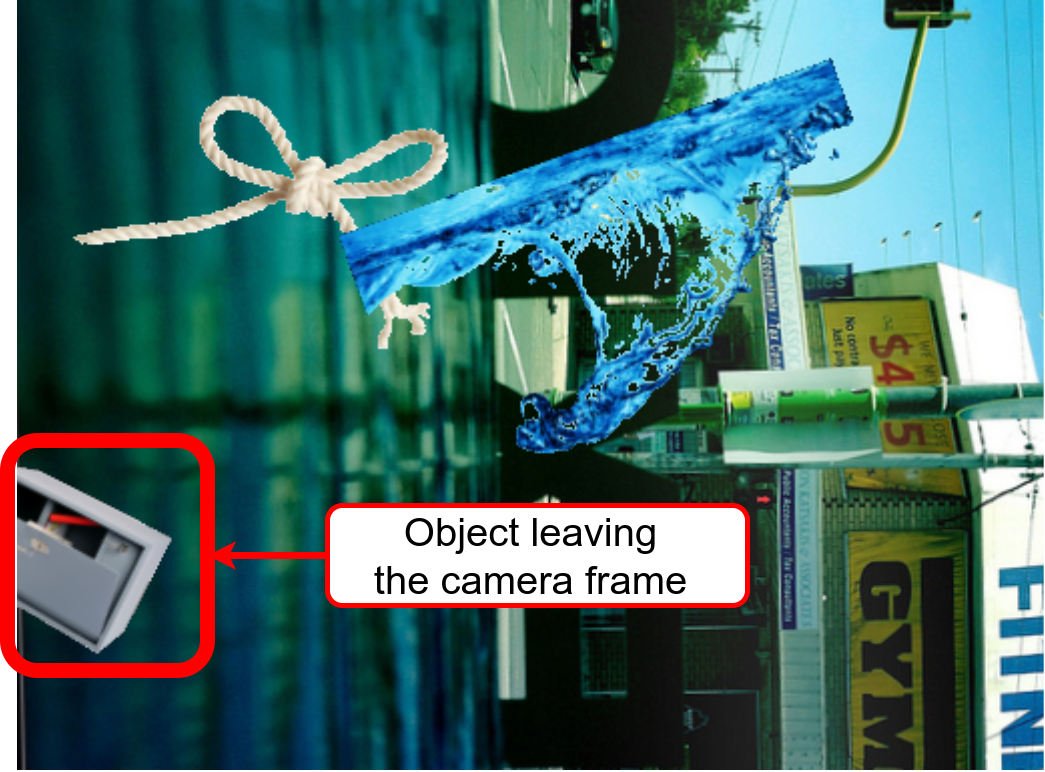} &
        \includegraphics[width=0.825\linewidth]{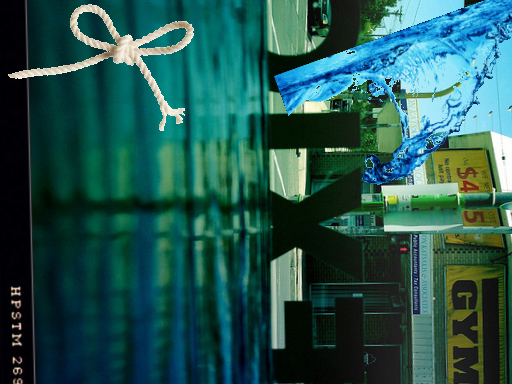} &
        \includegraphics[width=\linewidth]{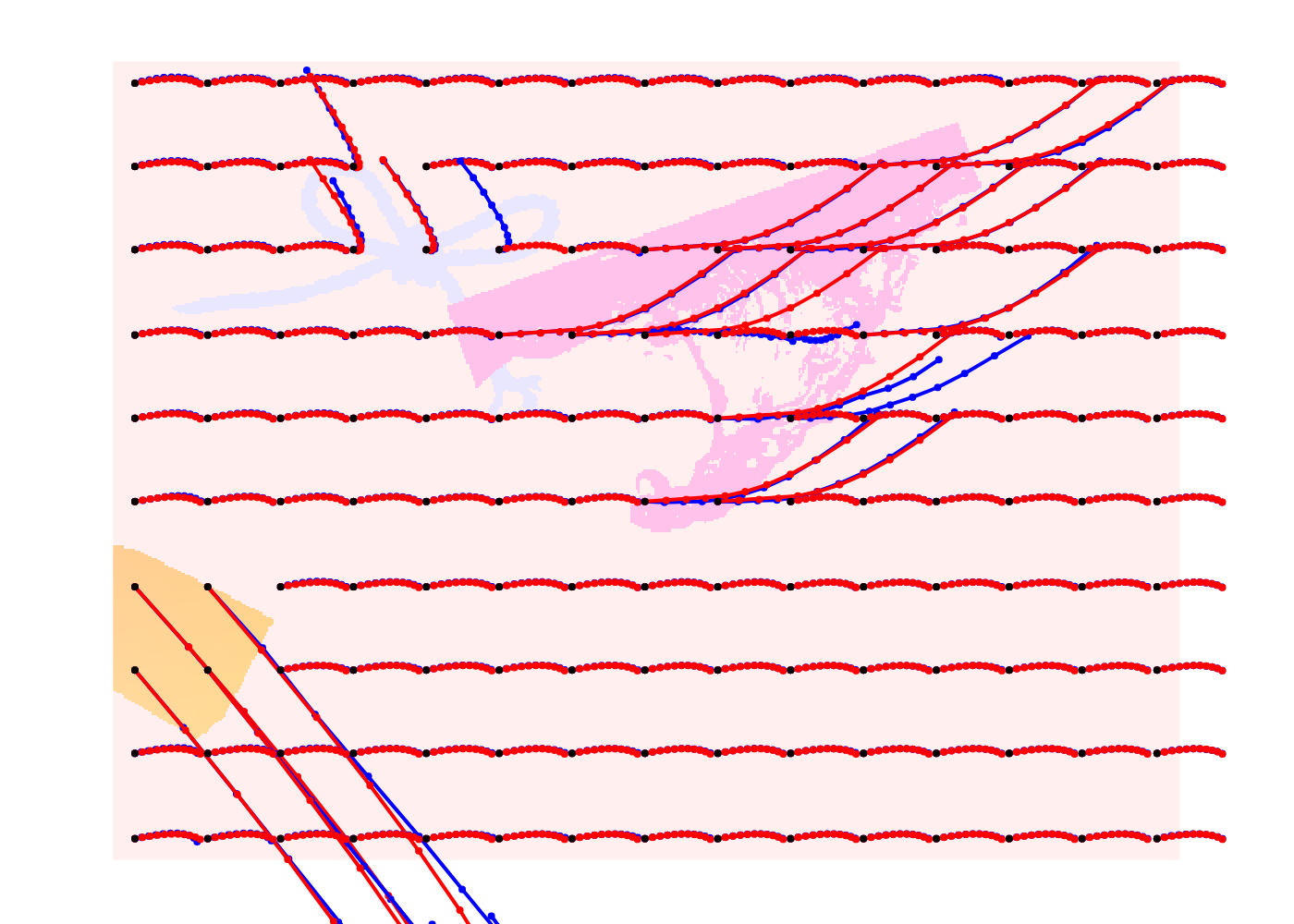} &
        \includegraphics[width=\linewidth]{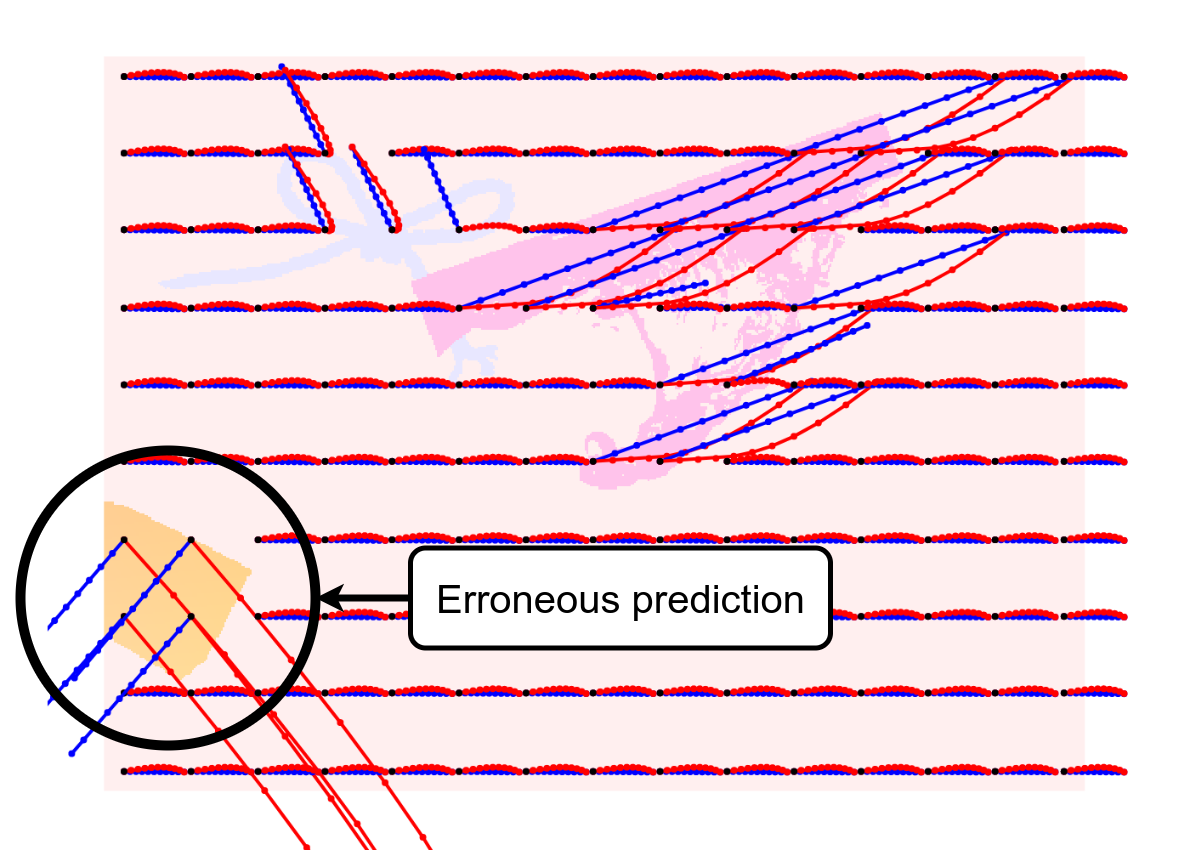}\\
        \includegraphics[width=0.825\linewidth]{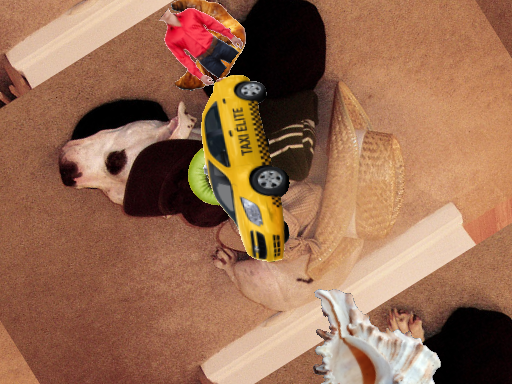} &
        \includegraphics[width=0.825\linewidth]{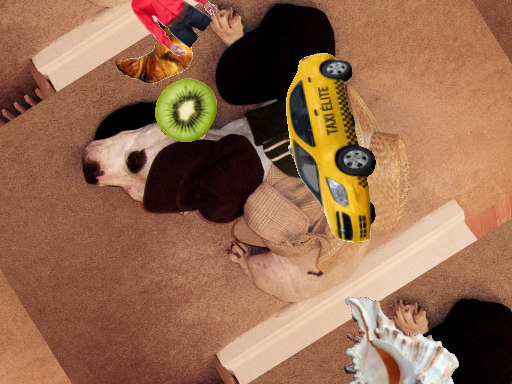} &
        \includegraphics[width=\linewidth]{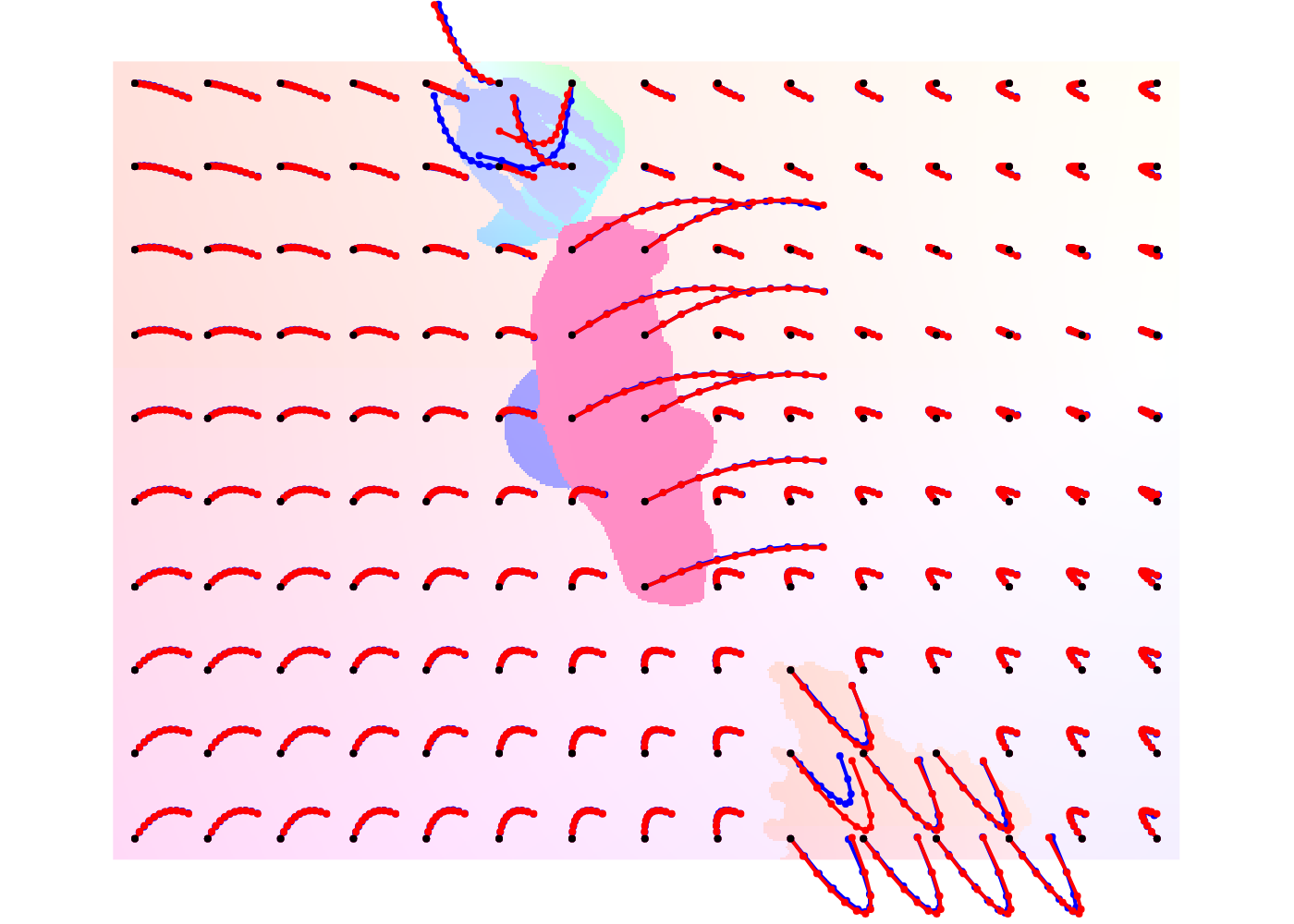} &
        \includegraphics[width=\linewidth]{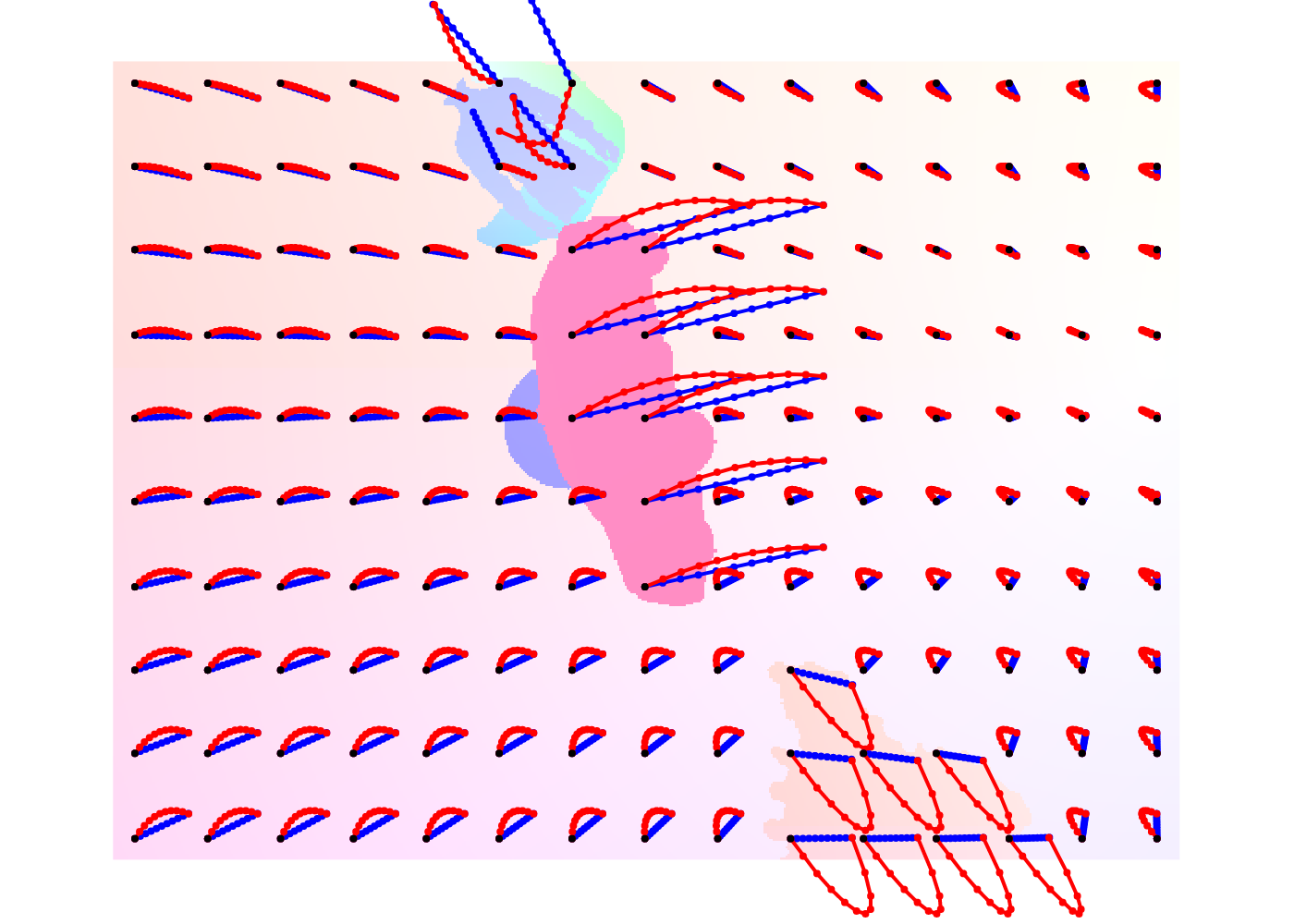}\\
        \includegraphics[width=0.825\linewidth]{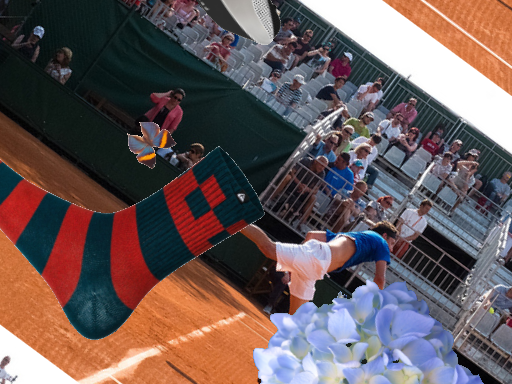} &
        \includegraphics[width=0.825\linewidth]{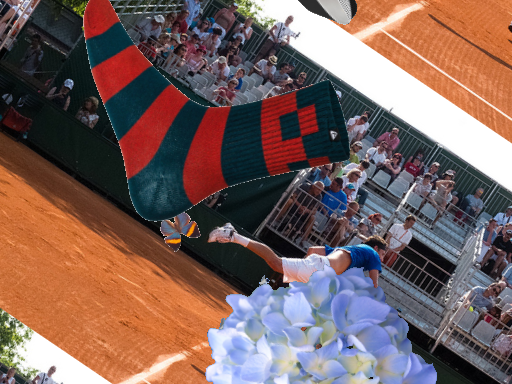} &
        \includegraphics[width=\linewidth]{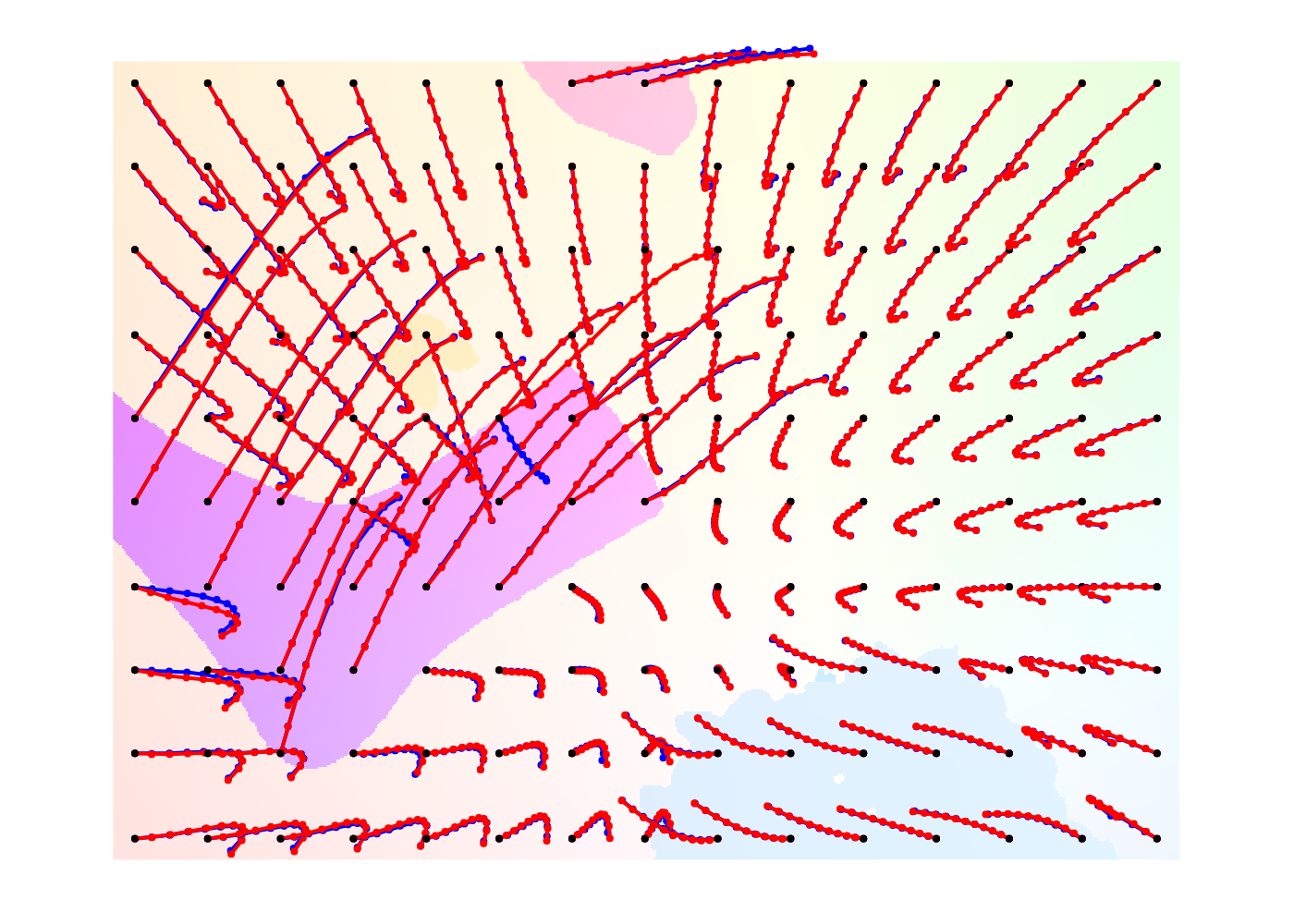} &
        \includegraphics[width=\linewidth]{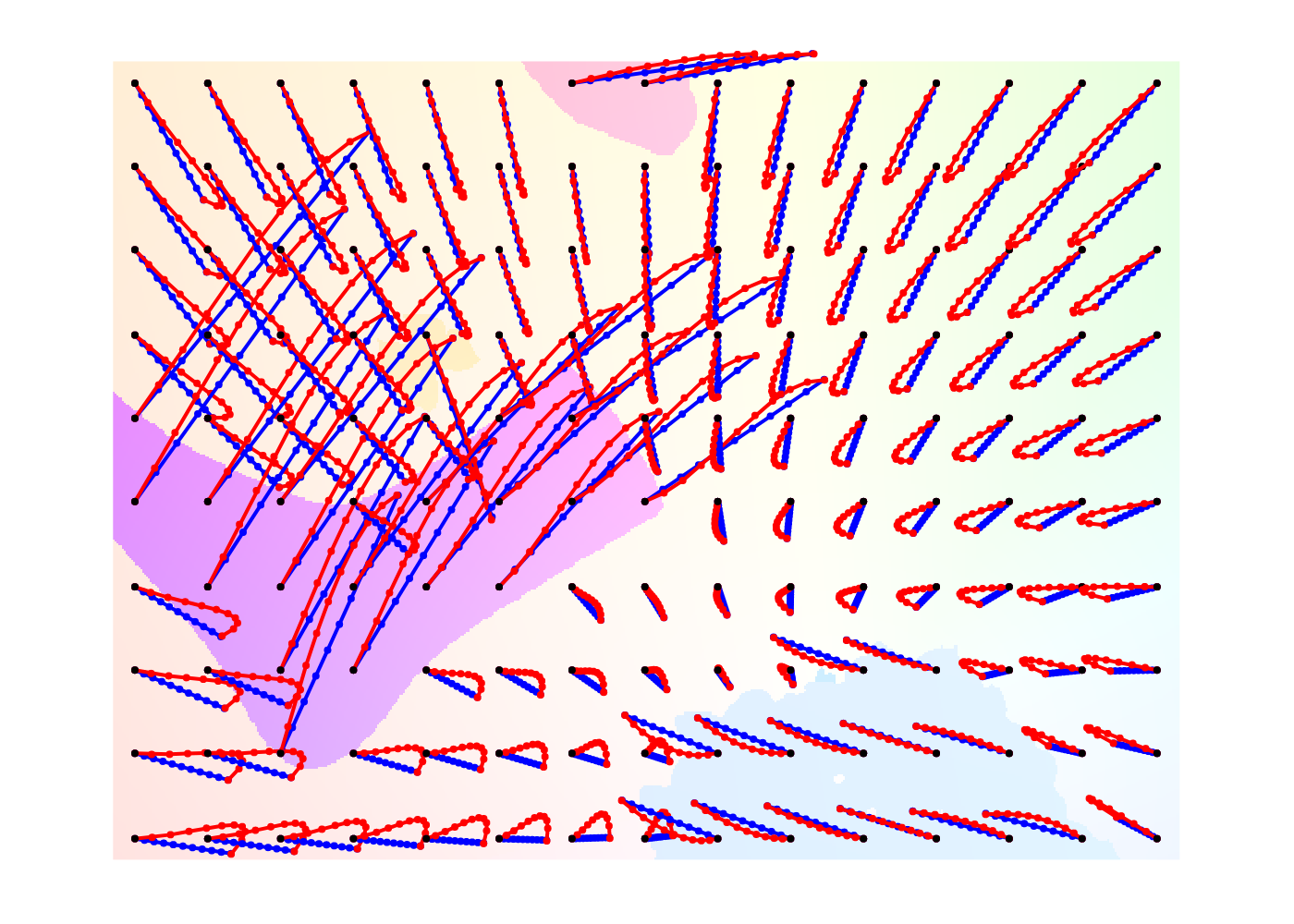}\\
        (a) First Frame & (b) Second Frame & (c) \textbf{Ours} & (d) {\scriptsize RAFT+GMA \cite{Jiang_2021_ICCV}}\\
    \end{tabular}
    \caption{Predictions of our method (c) and the strongest baseline, RAFT+GMA \cite{Jiang_2021_ICCV} (d), on the \simdset{} dataset. Predictions are shown in blue and the ground truth trajectory is visualized in red. The background is a colorization of the ground truth flow to highlight moving objects. Events are not shown for conciseness but are also used by our method. Best viewed in PDF form.}
    \label{fig:multiflow_examples}
\end{figure*}

%% file: tables/dsec_flow_exp.tex
\begin{table}[t!]
    \centering
\begin{adjustbox}{max width=\linewidth}

\begin{tabular}{lllllll}
\toprule
           & Input & EPE  & AE         & 1PE   & 2PE  & 3PE  \\ \midrule
EV-FlowNet \cite{Zhu18rss} & E     & 2.32 & -          & 55.4  & 29.8 & 18.6 \\
E-RAFT \cite{Gehrig3dv2021}    & E     & 0.79 & 2.85       & 12.74 & 4.74 & 2.68 \\
RAFT \cite{teed20eccv}       & I     & 0.78 & {\ul 2.44} & 12.40 & 4.60 & 2.61 \\
RAFT + GMA \cite{Jiang_2021_ICCV} & I & 0.94 & 2.66 & 12.98 & 5.08 & 2.96\\
\textbf{Ours} & E   & {\ul 0.75}    & 2.68          & {\ul 11.90}   & {\ul 4.41}    & {\ul 2.44}    \\
\textbf{Ours} & E+I & \textbf{0.69} & \textbf{2.42} & \textbf{9.70} & \textbf{3.42} & \textbf{1.88} \\\bottomrule%
\end{tabular}

\end{adjustbox}
\medskip
\caption{Results on DSEC-Flow. Our method outperforms prior work even when only using events.
Adding image data further improves performance.%
}%
\label{tab:dsec_flow_exp}
\end{table}

%% file: tables/multiflow_exp.tex
\begin{table}[t!]
\centering
\begin{adjustbox}{max width=\linewidth}

\begin{tabular}{@{}llllll@{}}
    \toprule
                  &       & \multicolumn{2}{l}{{\ul Trajectory Metrics}} & \multicolumn{2}{l}{{\ul 2-View Metrics}} \\
                  & Input & TEPE                & TAE                & EPE                   & AE                   \\ \midrule
    RAFT \cite{teed20eccv}         & I     & (6.89)                & (19.31)              & 7.42                  & 6.71                 \\
    RAFT + GMA \cite{Jiang_2021_ICCV} & I & (5.14) & (16.35) & \textbf{1.47} & \textbf{1.56}\\
    E-RAFT \cite{Gehrig3dv2021}        & E     & (6.70)                & (18.44)              & 7.56                  & 6.19                 \\
    \changes{E-RAFT + \bez{}} & \changes{E} & \changes{2.62} & \changes{5.92} & \changes{4.54} & \changes{6.06} \\
    \textbf{Ours} & E     & {\ul 1.85}          & {\ul 4.61}         & 3.37            & 4.80           \\
    \textbf{Ours} & E+I   & \textbf{1.29}       & \textbf{3.35}      & {\ul 2.27}         & {\ul 3.19}        \\\bottomrule %
\end{tabular}

\end{adjustbox}
\medskip
\caption{Results on \simdset{}. TEPE and TAE are the trajectory versions of EPE and AE. \changes{Metrics in brackets are computed using a linear motion model, highlighting that these methods are not originally designed for accurate trajectory prediction. E-RAFT + \bez{} represents our baseline model consisting of E-RAFT that estimates \bez{} curves instead of pixel displacements.}
}
\label{tab:multiflow_exp}
\end{table}

%% file: figs/timelens_examples_v2.tex
\begin{figure*}[!t]
    \centering
    \newcommand{\thisfigWidth}{0.193\linewidth}
    \newcommand{\boxwidth}{1.00\linewidth}
    \setlength\tabcolsep{1.5pt} %
    \begin{tabular}{M{\thisfigWidth}M{\thisfigWidth}M{\thisfigWidth}M{\thisfigWidth}M{\thisfigWidth}}
        \includegraphics[height=95pt]{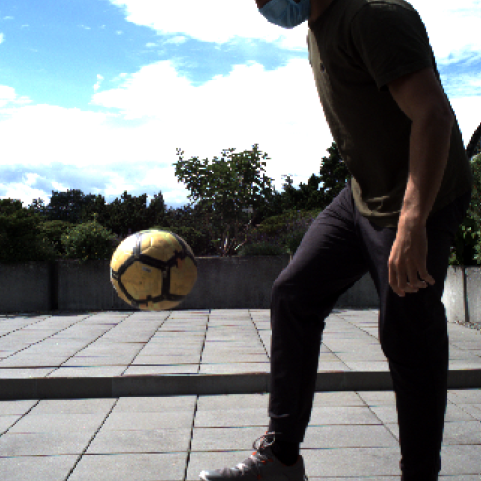} &
        \includegraphics[height=95pt]{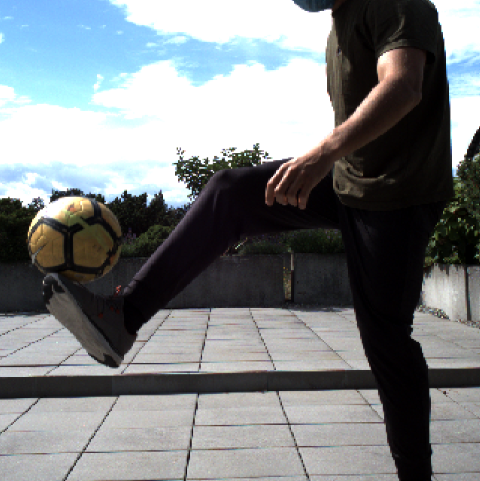} &
        \includegraphics[height=95pt]{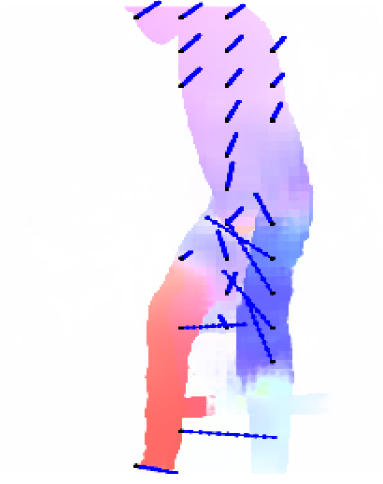} &
        \includegraphics[height=95pt]{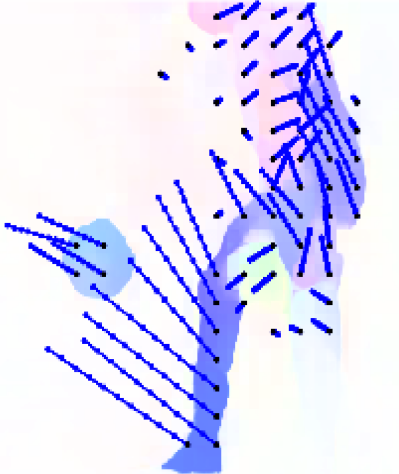} &
        \includegraphics[height=95pt]{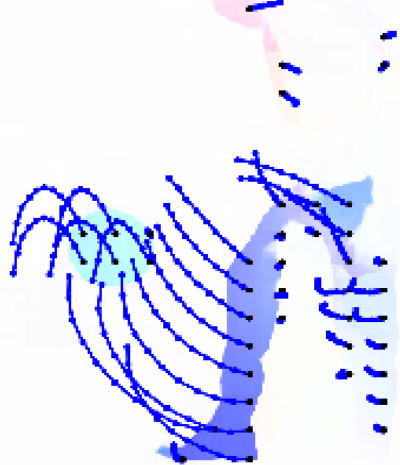} \\
        \includegraphics[height=95pt]{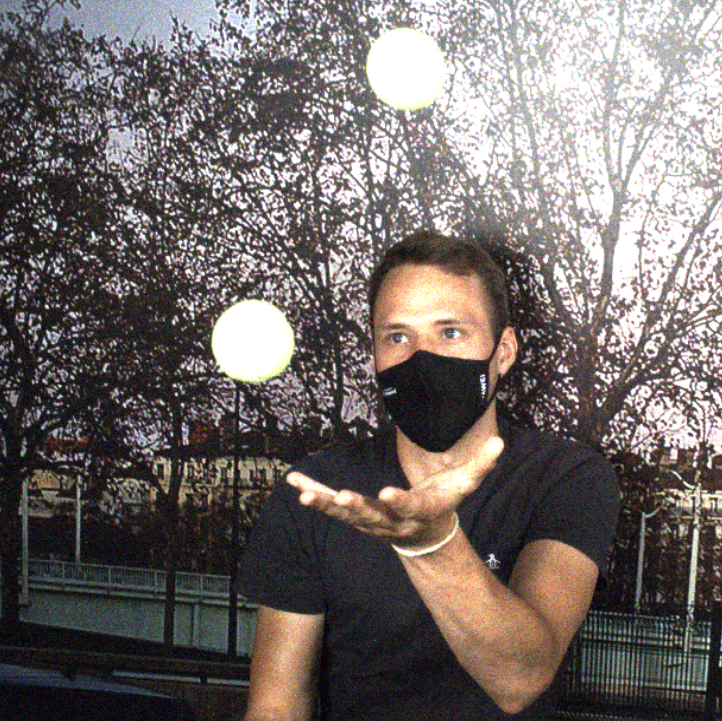} &
        \includegraphics[height=95pt]{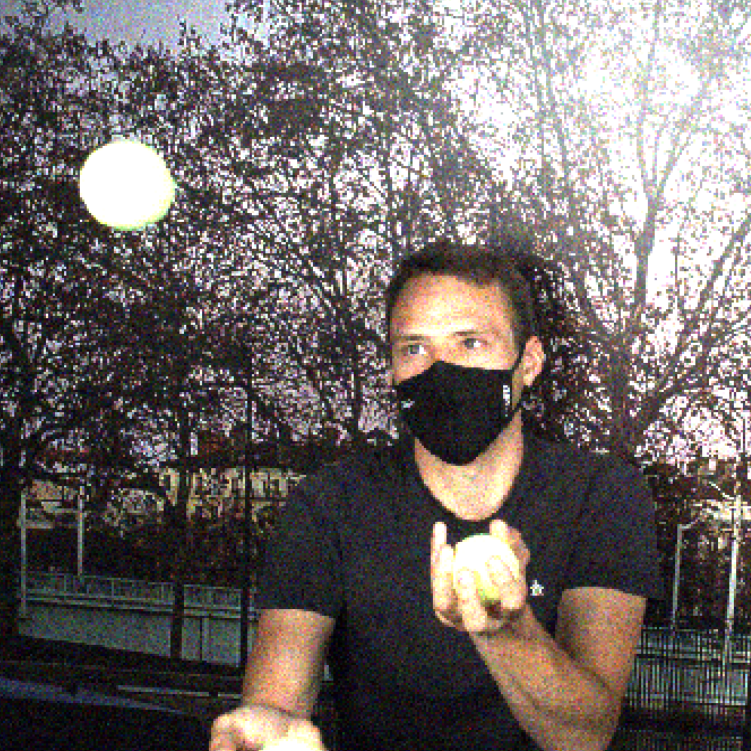} &
        \includegraphics[height=95pt]{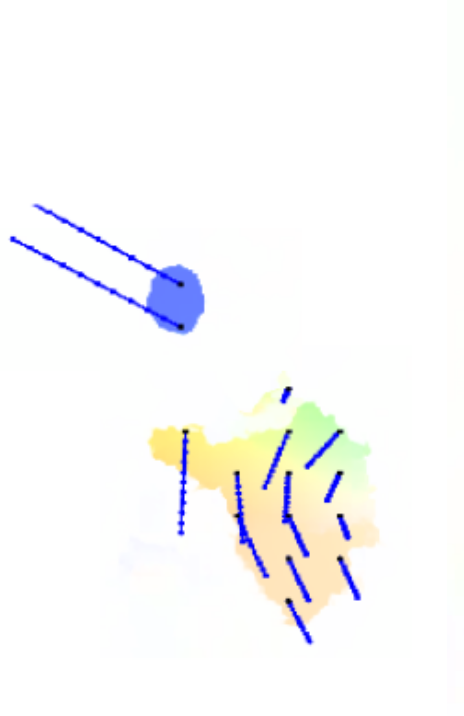} &
        \includegraphics[height=95pt]{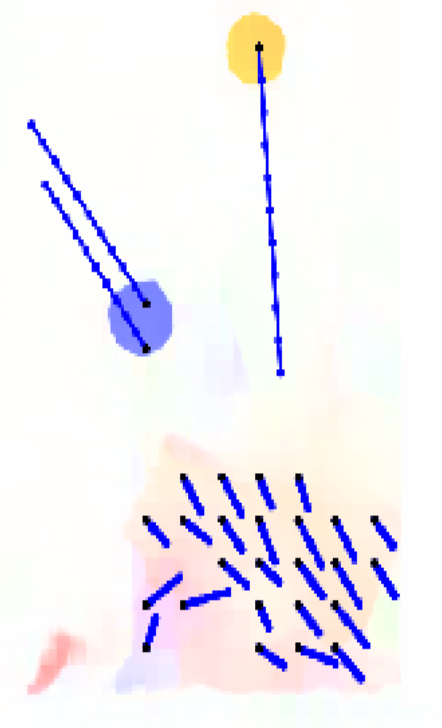} &
        \includegraphics[height=95pt]{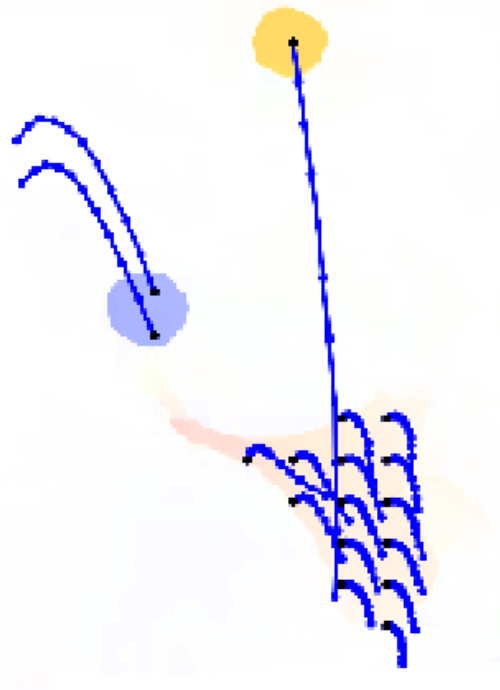} \\
        (a) First Frame & (b) Second Frame & (c) {\scriptsize RAFT+GMA \cite{Jiang_2021_ICCV}} & (d) \changes{E-RAFT \cite{Gehrig3dv2021}} & (e) \textbf{Ours}\\
    \end{tabular}
    \caption{\changes{Predictions of our method using only event data (e), E-RAFT \cite{Gehrig3dv2021} using events (d), and RAFT+GMA \cite{Jiang_2021_ICCV} using frames (c), on the \tldata{} \cite{Tulyakov21CVPR} dataset. Predictions are shown in blue. The background is a colorization flow to highlight moving objects. Events are not shown for conciseness.}}
    \label{fig:timelense_examples}
\end{figure*}

%% file: tables/ablation_exp.tex
\begin{table}[ht!]
    \centering
    \begin{adjustbox}{max width=\linewidth}
        \begin{tabular}{llllll}
            \toprule
                                                                      &       & \multicolumn{2}{l}{{\ul Trajectory Metrics}} & \multicolumn{2}{l}{{\ul 2-View Metrics}} \\
                                                                      & Input & TEPE                  & TAE                  & EPE                 & AE                 \\ \midrule
            w/o \bez{}                                                & E+I   & 6.36                  & 18.37                & 6.66                & 5.18               \\
            w/o multi-view                                            & E+I   & 1.81                  & 4.25                 & 2.68                & 3.51               \\
            {\ul Reference}                                           & E+I   & {\ul 1.29}            & {\ul 3.35}           & {\ul 2.27}          & {\ul 3.19}         \\ \midrule
            w/o \bez{}                                                & E     & 7.31                  & 20.07                & 9.10                & 8.85              \\
            w/o multi-view                                     & E     & 2.62                  & 5.92                 & 4.54                & 6.06               \\
            {\ul Reference}                                           & E     & {\ul 1.85}            & {\ul 4.61}           & {\ul 3.37}          & {\ul 4.80}         \\ \bottomrule
        \end{tabular}

    \end{adjustbox}
    \medskip
    \caption{Ablation experiments on \simdset{}. Final model settings are underlined.}
    \label{tab:ablation_exp}
\end{table}

%% file: figs/bez_degree.tex
\begin{figure}[!t]
    \centering
    \newcommand{\thisfigWidth}{0.45\linewidth}
    \newcommand{\boxwidth}{1.00\linewidth}
    \begin{tabular}{M{\thisfigWidth}M{\thisfigWidth}}
        \includegraphics[height=22ex]{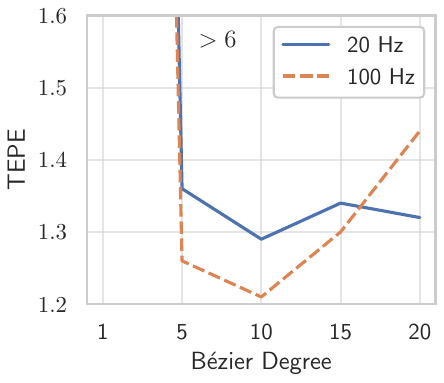} &
        \includegraphics[height=22ex]{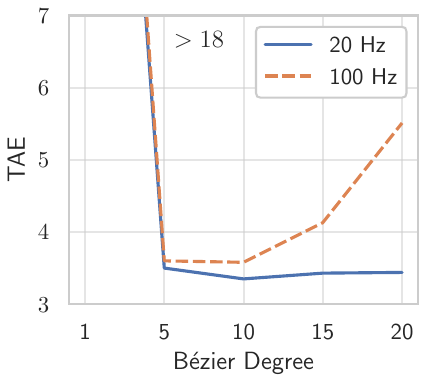}\\
        {\footnotesize (a) TEPE vs \bez{} degree} & {\footnotesize (b) TAE vs \bez{} degree} \\
    \end{tabular}
    \caption{\textbf{Trajectory metrics vs \bez{} curve degree}. (a) shows the trajectory EPE and (b) the trajectory AE as a function of the \bez{} degree. We evaluate both at 20 Hz (every 50 milliseconds) and at 100 Hz (every 10 ms). In both cases we can observe overfitting of the predicted trajectories when we evaluate at 100 Hz. This can be explained with the fact that the loss function was only applied every 50 milliseconds during training.}
    \label{fig:bez_degree}
\end{figure}

%% file: tables/supp/ablation_loss.tex
\begin{table}[ht!]
  \centering
  \begin{adjustbox}{max width=\linewidth}
  \begin{tabular}{@{}llllll@{}}
  \toprule
                        &       & \multicolumn{2}{l}{{\ul Trajectory Metrics}} & \multicolumn{2}{l}{{\ul 2-View Metrics}} \\
                        & Input & TEPE                  & TAE                  & EPE                 & AE                 \\ \midrule
  2-View Loss           & E+I   & 16.47                 & 20.57                & 3.72                & 5.48               \\
  {\ul Trajectory Loss} & E+I   & {\ul 1.29}            & {\ul 3.35}           & {\ul 2.27}          & {\ul 3.19}         \\ \midrule
  2-View Loss           & E     & 16.99                 & 21.83                & 5.95                & 8.57               \\
  {\ul Trajectory Loss} & E     & {\ul 1.85}            & {\ul 4.61}           & {\ul 3.37}          & {\ul 4.80}         \\ \bottomrule
  \end{tabular}
  \end{adjustbox}
  \medskip
  \caption{The trajectory loss refers to the loss function of equation (10) of the main paper, with $N_k>1$. The 2-View loss refers to the same loss function with $N_k=1$. Training our method with the trajectory loss leads to drastically lower errors even for the 2-View metrics.}
  \label{tab:ablation_loss}
\end{table}

%% file: tables/supp/ablation_num_corr.tex
\begin{table}[ht!]
  \centering

  \begin{tabular}{@{}cllll@{}}
  \toprule
  & \multicolumn{2}{l}{{\ul Trajectory Metrics}} & \multicolumn{2}{l}{{\ul 2-View Metrics}} \\
   \# Correlation Lookups & TEPE                  & TAE                  & EPE                 & AE                 \\ \midrule
1           & 1.81                 & 4.25                & 2.68          & 3.51               \\
3           & 1.35                 & 3.43                & 2.34          & 3.21               \\
{\ul 5}           & {\ul 1.29}                 & {\ul 3.35}                & {\ul 2.27}          & {\ul 3.19}               \\\bottomrule
  \end{tabular}
  \medskip
  \caption{Increasing the number of correlation volumes/lookups in time improves performance. The underlined row refers to the reference model used in the experiments.}
  \label{tab:ablation_num_corr}
\end{table}

%% file: tables/supp/ablation_voxel_bins.tex
\begin{table}[ht!]
  \centering

\begin{changedtable}
  \begin{tabular}{@{}cllll@{}}
  \toprule
  & \multicolumn{2}{l}{{\ul Trajectory Metrics}} & \multicolumn{2}{l}{{\ul 2-View Metrics}} \\
   Bin Size & TEPE                  & TAE                  & EPE                 & AE                 \\ \midrule
0.1           & 1.81                  & 4.53                 & 2.87           & 3.97                \\
0.05           & 1.51                 & 3.82                 & 2.61           & 3.63                \\
0.025           & 1.40                 & 3.56                 & 2.45          & 3.41               \\
{\ul 0.0125}           & {\ul 1.29}                 & {\ul 3.35}                & {\ul 2.27}          & {\ul 3.19}               \\\bottomrule
  \end{tabular}
  \medskip
  \caption{Decreasing the bin size of the voxel grid improves performance. Each bin is assigned a specific timestamp and the bin size refers to the $\Delta t$ between the bins. The underlined row refers to the reference model used in the experiments.}

\end{changedtable}
  \label{tab:ablation_voxel_bins}
\end{table}

%% file: tables/supp/model_stats.tex
{
\setlength{\tabcolsep}{4pt}
\newcommand{\raftParams}{5.3}
\newcommand{\eraftParams}{5.3}
\newcommand{\eParams}{5.6}
\newcommand{\eiParams}{5.9}
\newcommand{\raftTime}{52}
\newcommand{\eraftTime}{56}
\newcommand{\eTime}{77}
\newcommand{\eiTime}{84}
\newcommand{\raftMemory}{1.20}
\newcommand{\eraftMemory}{1.20}
\newcommand{\eMemory}{1.65}
\newcommand{\eiMemory}{1.65}

\begin{table}[ht!]
    \centering
    \begin{tabular}{@{}lllll@{}}\toprule
        & Input & Inference time {[}ms{]} & Params & Memory {[}GB{]}\\ \midrule
    RAFT \cite{teed20eccv} & I    & \raftTime               & \raftParams{} M  & \raftMemory           \\
    E-RAFT \cite{Gehrig3dv2021} & E    & \eraftTime               & \eraftParams{} M  & \eraftMemory           \\
    Ours & E    & \eTime               & \eParams{} M  & \eMemory           \\
    Ours & E+I  & \eiTime               & \eiParams{} M  & \eiMemory           \\\bottomrule
    \end{tabular}
    \medskip
    \caption{Comparison of inference time, parameter count and memory consumption on \simdset{}. These numbers have been obtained on a Titan RTX GPU with an implementation using Pytorch version \href{https://github.com/pytorch/pytorch/releases/tag/v1.12.1}{1.12.1}}
    \label{tab:model_stats}
\end{table}
}

%% file: sections/discussion.tex
\section{Limitations}
The current implementation computes the dot products between all paired features at once, which incurs $\mathcal{O}(N^2)$ space complexity where N is the number of spatial features.
This limitation can be overcome by computing dot products on-demand, as original proposed by Teed et al. \cite{teed20eccv}.
Another limitation of the proposed method is that we cannot compute the pixel trajectories for a sequence longer than the current input to the model.
We can approximate longer trajectories by concatenating multiple shorter sequences via interpolation at the reference timestamps.
A more elegant solution could be achieved by designing a recurrent model that can track pixels without assuming a fixed sequence length or duration.
Finally, the number of correlation lookups is a fixed parameter that cannot be changed at inference time.
As a result, we must define at training time how many correlation lookups can be afforded during inference time.

\section{Conclusion}
We have introduced a method for estimating continuous-time pixel trajectories from events and frames.
The proposed approach generalizes the frame-based \cite{teed20eccv} and event-based \cite{Gehrig3dv2021} RAFT architecture by regressing \bez{} curves and uses the pixel trajectories to extract correlation features along the temporal axis.
Our experimental results demonstrate that the proposed method can accurately predict continuous pixel trajectories while at the same time outperforming strong baselines not only in simulation but also on real data. 
Finally, our sim2real results suggest that the new \simdset{} dataset can also be used to pretrain pixel trajectory regression models for downstream applications.